\documentclass{article}

\usepackage{url}

\usepackage{hyperref}% http://ctan.org/pkg/hyperref
\hypersetup{
  colorlinks=true,
	citecolor=blue!50!blue,
  linkcolor=blue!50!blue,
  urlcolor=blue!70!blue
}

\usepackage{times}
\usepackage{rotating}
\usepackage{multirow}
\usepackage{natbib}
\usepackage{amssymb}
\usepackage{amsmath}
\usepackage{mathtools}
\usepackage{multirow}
\usepackage{tcolorbox}
\usepackage{tabularx}
\usepackage{array}
\usepackage{colortbl}
\usepackage{fancybox}
\usepackage{tikz}

\tcbuselibrary{skins}

\usepackage{pifont}% http://ctan.org/pkg/pifont
\usepackage[T1]{fontenc}    % use 8-bit T1 fonts
\usepackage{hyperref}       % hyperlinks
\usepackage{url}            % simple URL typesetting

\usepackage{amsfonts}       % blackboard math symbols
\usepackage{nicefrac}       % compact symbols for 1/2, etc.
\usepackage{microtype}      % microtypography
\usepackage{amssymb}
\usepackage{amsmath}
\usepackage{mathtools}
\usepackage{bigdelim}

\usepackage{color}
\usepackage{verbatim}
\usepackage{siunitx}

\tcbuselibrary{skins}
\usepackage{fancybox}
\usepackage{tikz}
\usepackage{multirow}
\usepackage{tcolorbox}
\usepackage{tabularx}
\usepackage{array}
\usepackage{colortbl}
\usepackage{epstopdf}

\usepackage{xcolor,colortbl}

\definecolor{Gray}{gray}{0.85}
\definecolor{LightCyan}{rgb}{0.88,1,1}

\newcolumntype{a}{>{\columncolor{Gray}}c}
\newcolumntype{b}{>{\columncolor{white}}c}

\usepackage{algorithm}
\usepackage{hyperref}

\tcbuselibrary{skins}

\tcbset{tab3/.style={fonttitle=\bfseries\footnotesize,fontupper=\footnotesize,
colback=yellow!5!white,colframe=black!75!white,colbacktitle=white!40!white,
coltitle=black,center title,freelance}}

\usepackage{tikz}
\usetikzlibrary{positioning}

\usepackage[font=small,skip=5.5pt]{caption}

\usepackage[utf8]{inputenc} % allow utf-8 input
\usepackage[T1]{fontenc}    % use 8-bit T1 fonts
\usepackage{hyperref}       % hyperlinks
\usepackage{url}            % simple URL typesetting
\usepackage{booktabs}       % professional-quality tables
\usepackage{amsfonts}       % blackboard math symbols
\usepackage{nicefrac}       % compact symbols for 1/2, etc.
\usepackage{microtype}      % microtypography

\usepackage{color}
\usepackage{enumitem}

\usepackage{calc}

\newlength{\spacer}
\newsavebox{\mybox}

\usepackage{microtype}
\usepackage{graphicx}
\usepackage{subfigure}
\usepackage{booktabs} % for professional tables

\usepackage{hyperref}

\usepackage[accepted]{icml2020}

\icmltitlerunning{Quantifying Uncertainty in Deep Learning via Higher-Order Influence Functions}

\begin{document}

\twocolumn[
\icmltitle{Discriminative Jackknife: Quantifying Uncertainty in Deep Learning via Higher-Order Influence Functions}




\begin{icmlauthorlist}
\icmlauthor{Ahmed M. Alaa}{to}
\icmlauthor{Mihaela van der Schaar}{to,too}
\end{icmlauthorlist}

\icmlaffiliation{to}{UCLA}
\icmlaffiliation{too}{Cambridge University}

\icmlcorrespondingauthor{Ahmed M. Alaa}{ahmedmalaa@ucla.edu}

\icmlkeywords{Machine Learning, ICML}

\vskip 0.3in
]



\printAffiliationsAndNotice{}  % leave blank if no need to mention equal contribution

\begin{abstract}
Deep learning models achieve high predictive accuracy across a broad spectrum of tasks, but rigorously quantifying their predictive~uncertainty~remains challenging. Usable estimates of predictive uncertainty should (1) {\it cover} the true prediction targets with high probability,~and~(2)~{\it discriminate} between high- and low-confidence prediction instances. Existing methods~for~uncertainty quantification are based predominantly~on~Bayesian~neural networks; these may fall short of (1) and (2) --- i.e., Bayesian credible intervals do not guarantee frequentist coverage, and approximate posterior inference undermines discriminative accuracy. In this paper, we develop the~{\it discriminative~jackknife} (DJ), a frequentist procedure that~utilizes~influence functions of a model's loss functional to construct a jackknife (or leave-one-out) estimator of predictive confidence intervals. The DJ satisfies (1) and (2), is applicable to a~wide~range~of deep learning models, is easy to implement, and can be applied in a {\it post-hoc} fashion without interfering with model training or compromising its accuracy. Experiments demonstrate that DJ performs competitively compared to existing Bayesian and non-Bayesian regression baselines.
\end{abstract}

\section{Introduction}
\label{Sec1}
Deep learning models have achieved state-of-the-art performance on a variety of learning tasks, and are becoming increasingly popular in various application domains \cite{lecun2015deep}. A key question often asked of such models is ``{\it Can we trust this particular model prediction?}''~This~question is highly relevant in high-stakes applications wherein predictions are used to inform critical decision-making --- examples of such applications include: medical decision support \citep{alaa2017bayesian}, autonomous vehicles, and financial forecasts \cite{amodei2016concrete}.~Despite~their~impressive~accuracy, rigorously quantifying uncertainty in deep learning models is a challenging and yet an unresolved problem \cite{gal2016uncertainty, ovadia2019can}. 

\begin{figure*}[t]
  \centering
  \includegraphics[width=6.25in]{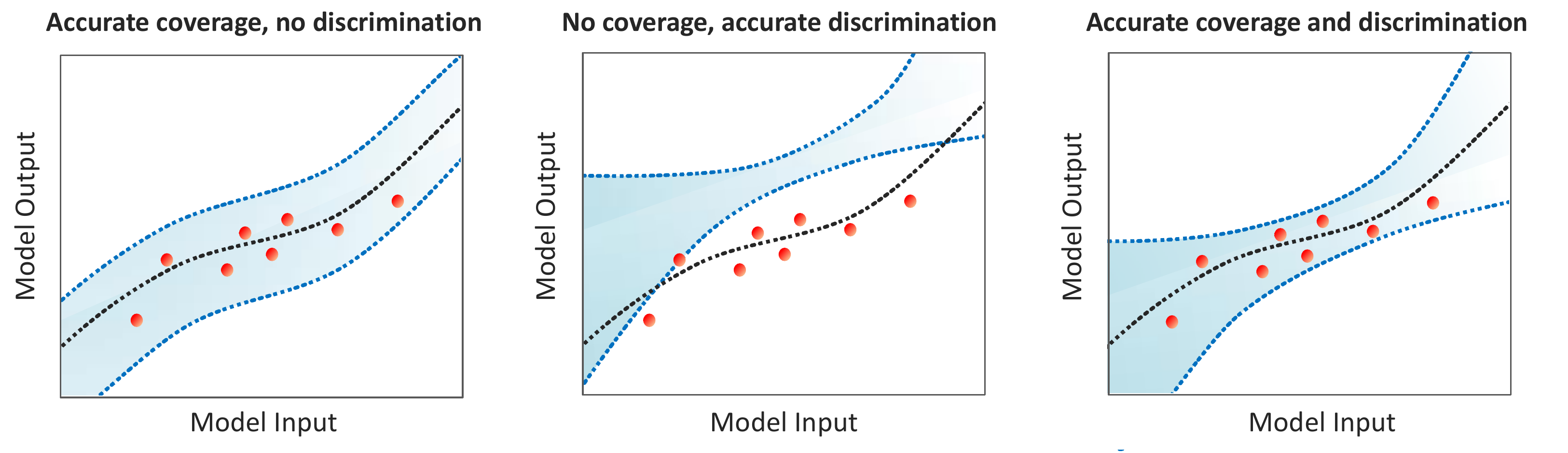}
  \caption{\footnotesize {\bf Pictorial depiction for coverage and discrimination in uncertainty estimates.} Red dots correspond to training data and~dotted black line corresponds to the target function. Confidence intervals are visualized as shaded blue regions, where dotted~blue~lines~are~the upper and lower confidence limits. The left panel shows a confidence interval that perfectly covers the data points, but does not discriminate high-confidence predictions (regions with dense training data) and low-confidence ones (regions with scarce training data). The middle panel shows a confidence interval with a width proportional to the density of training data (which determines model uncertainty), but does not cover any data point. The right panel shows a confidence interval that satisfies both coverage and discrimination requirements.} 
  \rule{\linewidth}{.75pt}    
  \label{Fig1} 
  \vspace{-6mm} 
\end{figure*}

Actionable estimates of predictive uncertainty are ones that (1) {\it cover} the true prediction targets~with a high~probability, and (2) {\it discriminate} between high- and low-confidence predictions. (Figure \ref{Fig1} depicts a pictorial visualization for these coverage and discrimination requirements.) The coverage requirement is especially relevant in applications where predictive uncertainty is incorporated in a decision-theoretic framework (e.g., administering medical treatments \cite{dusenberry2019analyzing}, or estimating value functions in model-free reinforcement learning \cite{white2010interval}). The second requirement, discrimination, is crucial for auditing model reliability \cite{schulam2019can}, detecting dataset shifts and out-of-distribution samples \cite{barber2019conformal}, and actively collecting new training examples for which the model is not confident \cite{cohn1996active}. 

Existing methods for uncertainty estimation are based predominantly on Bayesian neural networks~(BNNs),~whereby predictive uncertainty is evaluated via posterior credible intervals \cite{welling2011bayesian,hernandez2015probabilistic, ritter2018scalable, maddox2019simple}. However, BNNs require significant modifications to the training procedure, and exact Bayesian~inference~is computationally~prohibitive in practice. Approximate dropout-based inference schemes (e.g., Monte Carlo dropout \cite{gal2016dropout} and variational dropout \cite{kingma2015variational}) have been recently proposed as computationally efficient alternatives. However, Bayesian inference in dropout-based models has been shown to be ill-posed, since the induced posterior distributions in such models do not concentrate asymptotically \cite{osband2016risk, hron2017variational}, which jeopardizes both the coverage and discrimination performance of the resulting credible intervals. Moreover, even with exact inference, Bayesian credible intervals do not guarantee frequentist coverage \cite{bayarri2004interplay}. Non-Bayesian alternatives have been recently developed based on ad-hoc ensemble designs \cite{lakshminarayanan2017simple} --- but formal and rigorous frequentist methods are still lacking.

{\bf Summary of contributions.} In this paper, we develop a formal procedure for constructing frequentist (pointwise) {\it confidence intervals} on the predictions of a broad class of deep learning models. Our method --- which we call the {\it discriminative jackknife} (DJ) --- is inspired by the classic jackknife leave-one-out (LOO) re-sampling procedure for estimating variability in statistical models \cite{miller1974jackknife,efron1992jackknife}. In order to ensure both frequentist coverage and discrimination, DJ constructs feature-dependent confidence intervals using the LOO local prediction variance at the input feature, and adjusts the interval width (for a given coverage probability) using the model's average LOO error residuals. Whereas the classic jackknife satisfies~neither~the coverage nor the discrimination requirements \cite{barber2019predictive}, DJ satisfies both (i.e., DJ generates predictive confidence intervals resembling those in the rightmost panel of Figure \ref{Fig1} with high probability).

Central to our DJ procedure is the use of~{\it influence~functions} --- a key concept in robust~statistics~and variational~calculus \cite{cook1982residuals, efron1992jackknife} ---~in~order~to~estimate the parameters of models trained on LOO~versions~of the training data, without exhaustively re-training the model for each held-out data point. That is, using the {\it von Mises} expansion \cite{fernholz2012mises} --- a variant of Taylor series expansion for statistical functionals --- we represent the (counter-factual) model parameters that would have been learned on LOO versions of the training data set in terms of an infinite series of higher-order influence functions (HOIFs) for the model parameters trained on the complete data. To compute the second-order von Mises expansion, we derive an approximate formula for evaluating second-order~influence functions that extends on the formula for first-order influence in \cite{koh2017understanding}. We also propose a general procedure for computing HOIFs by {\it recursively} computing hessian-vector products between the Hessian and higher-order gradients of the model loss, without the need for explicitly inverting the Hessian matrix. 

Comprehensive experimental evaluation demonstrates that the DJ performs competitively compared to both Bayesian and non-Bayesian methods with respect to both the coverage and discrimination criteria. Because of the~{\it post-hoc}~nature of the DJ, it is capable of improving coverage and discrimination without any modifications to the underlying predictive model. However, since computing influence functions entails at least linear complexity in both the number of training data points and the number of model parameters, a key limitation of our method is scalability. We identify computationally efficient methods for approximating HOIFs as an interesting direction for future research.   

\section{Preliminaries}
\label{Sec2}
\subsection{Learning Setup}
\label{Sec21}
We consider a standard supervised learning setup with $(\boldsymbol{x}, y)$ being a feature-label pair, where the feature~$\boldsymbol{x}$~belongs~to~a $d$-dimensional feature space $\mathcal{X} \subseteq \mathbb{R}^d$, and $y \in \mathcal{Y}$. A model is trained to predict $y$ using a dataset $\mathcal{D}_n \triangleq \{(\boldsymbol{x}_i, y_i)\}^n_{i=1}$~of~$n$ examples, which are drawn i.i.d from a distribution $\mathbb{P}$. Let $f(\boldsymbol{x};\theta): \mathcal{X} \to \mathcal{Y}$ be the prediction model, where $\theta \in \Theta$ are the model parameters, and $\Theta$ is the parameter space. The trained parameters $\hat{\theta} \in \Theta$ are obtained by solving the optimization problem $\hat{\theta} = \arg \min_{\theta \in \Theta} L(\mathcal{D}_n, \theta)$, for a loss 
\begin{equation}
L(\mathcal{D}_n, \theta) \triangleq \frac{1}{n}\sum_{i=1}^{n}\,\ell(y_i, f(\boldsymbol{x}_i;\theta)),      
\label{eq1}
\end{equation}
where we fold in any regularization~terms~into~$\ell(.)$.~We~do not pose any assumptions on the specific architecture underlying the model $f(\boldsymbol{x};\theta)$; it~can~be~any~neural~network~variant, such as feed-forward or convolutional network. 

\subsection{Uncertainty Quantification}
\label{Sec22}
The predictions issued by the (trained) model are given by $f(\boldsymbol{x};\hat{\theta})$; our main goal is to obtain an estimate of uncertainty in the model's prediction, expressed through the pointwise confidence interval $\mathcal{C}(\boldsymbol{x};\hat{\theta})$, formally defined as follows: 
\begin{equation}
\mathcal{C}(\boldsymbol{x};\hat{\theta}) \, \triangleq \, [\,f_-(\boldsymbol{x};\hat{\theta}),\, f_+(\boldsymbol{x};\hat{\theta})\,], \, \forall \boldsymbol{x} \in \mathcal{X}.
\label{eq2}
\end{equation}
The degree of uncertainty in the model's prediction (for a data point with feature $\boldsymbol{x}$) is quantified by the {\it interval width} $W(.)$ of the confidence interval $\mathcal{C}(\boldsymbol{x};\hat{\theta})$, given by
\begin{align}
W(\mathcal{C}(\boldsymbol{x};\hat{\theta})) \, \triangleq \, f_+(\boldsymbol{x};\hat{\theta}) - f_-(\boldsymbol{x};\hat{\theta}).
\end{align}
Wider intervals imply less confidence, and vice versa. For $\mathcal{C}(\boldsymbol{x};\hat{\theta})$ to be usable, it has to satisfy the following:

(i) {\it Frequentist coverage.} This is satisfied if the confidence interval $\mathcal{C}(\boldsymbol{x};\hat{\theta})$ covers the true target $y$ with a prespecified coverage probability of $(1-\alpha),$ for $\alpha \in (0,1)$, i.e.,  
\begin{equation}
\mathbb{P}\left\{y \in \mathcal{C}(\boldsymbol{x};\hat{\theta})\right\} \geq 1 - \alpha, \nonumber  
\label{eq3}
\end{equation}
where the probability is taken with respect to a (new) test point $(\boldsymbol{x}, y)$ as well as with respect to the training data $\mathcal{D}_n$ \cite{lawless2005frequentist, barber2019predictive}. 

(ii) {\it Discrimination.} This requirement is met when $\mathcal{C}(\boldsymbol{x};\hat{\theta})$ is wider for test points with less accurate predictions \cite{leonard1992neural}, i.e., for the test points $\boldsymbol{x}, \boldsymbol{x}^{\prime} \in \mathcal{X}$, we have
\begin{align}
\mathbb{E}\big[\,W(\mathcal{C}(\boldsymbol{x};\hat{\theta}))\,\big] \, &\geq \, \mathbb{E}\big[\,W(\mathcal{C}(\boldsymbol{x}^{\prime};\hat{\theta}))\,\big] \,\, \Leftrightarrow \nonumber\\ 
\mathbb{E}\big[\,\ell(y, f(\boldsymbol{x};\hat{\theta}))\,\big] \, &\geq \, \mathbb{E}\big[\,\ell(y^{\prime}, f(\boldsymbol{x}^{\prime};\hat{\theta}))\,\big],\nonumber
\label{eq4}
\end{align}
where the expectation $\mathbb{E}[\,.\,]$ is taken with respect to the randomness of $\mathcal{D}_n$. In the next Section, we~develop~a~post-hoc frequentist procedure for estimating $\widehat{\mathcal{C}}(\boldsymbol{x};\hat{\theta})$ that satisfies both of the requirements in (i) and (ii).

\section{The Discriminative Jackknife} % Which Deep L models?
\label{Sec3}
Before presenting our {\it discriminative jackknife} (DJ) procedure, we start with a brief recap of the classical jackknife. The jackknife quantifies predictive uncertainty in terms of the (average) prediction error, which is estimated with a leave-one-out (LOO) construction found by systematically leaving out each sample in $\mathcal{D}_n$, and evaluating the error of the re-trained model on the held-out sample, i.e., for a target coverage of $(1-\alpha)$, the naïve jackknife is \cite{efron1992jackknife}:
\begin{equation}
\widehat{\mathcal{C}}^J_{\alpha}(\boldsymbol{x};\hat{\theta}) = f(\boldsymbol{x};\hat{\theta})\, \pm \, \widehat{Q}^+_{\alpha}(\mathcal{R}),
\label{eq5}
\end{equation}
with $\mathcal{R} = \left\{r_1,\ldots,r_n\right\}$, where $r_i = |\, y_i - f(\boldsymbol{x}_i;\hat{\theta}_{-i}) \,|$ is the error residual on the $i$-th data point, $\hat{\theta}_{-i}$ are the parameters of the model re-trained on the dataset~$\mathcal{D}_n \setminus \{(\boldsymbol{x}_i, y_i)\}$~(with the $i$-th point removed), and $\widehat{Q}^+_{\alpha}$ is the $(1 - \alpha)$ empirical quantile of the set $\mathcal{R} = \left\{r_1,\ldots,r_n\right\}$, defined as
\[\widehat{Q}^+_{\alpha}(\mathcal{R})\,\triangleq\,\mbox{the $\lceil (1-\alpha)\,(n + 1) \rceil$-th smallest value in $\mathcal{R}$},\]
where $\widehat{Q}^-_{\alpha}(\mathcal{R}) = \widehat{Q}^+_{1-\alpha}(-\mathcal{R})$. Albeit intuitive, the naïve jackknife is not guaranteed to achieve the target coverage \cite{barber2019predictive}. More crucially, the interval width $W(\widehat{\mathcal{C}}^J_{\alpha}(\boldsymbol{x};\hat{\theta}))$ is a constant (independent of $\boldsymbol{x}$), which renders discrimination impossible, i.e., naïve jackknife would result in intervals resembling the leftmost panel in Figure \ref{Fig1}.
%\footnote{It follows from this definition that $\widehat{Q}_{\alpha}(\mathcal{R}) = \widehat{Q}_{1-\alpha}(-\mathcal{R})$.}
\subsection{Exact Construction of the DJ Confidence Intervals}
\label{Sec31}
\begin{figure}[t]
  \centering
  \includegraphics[width=3.25in]{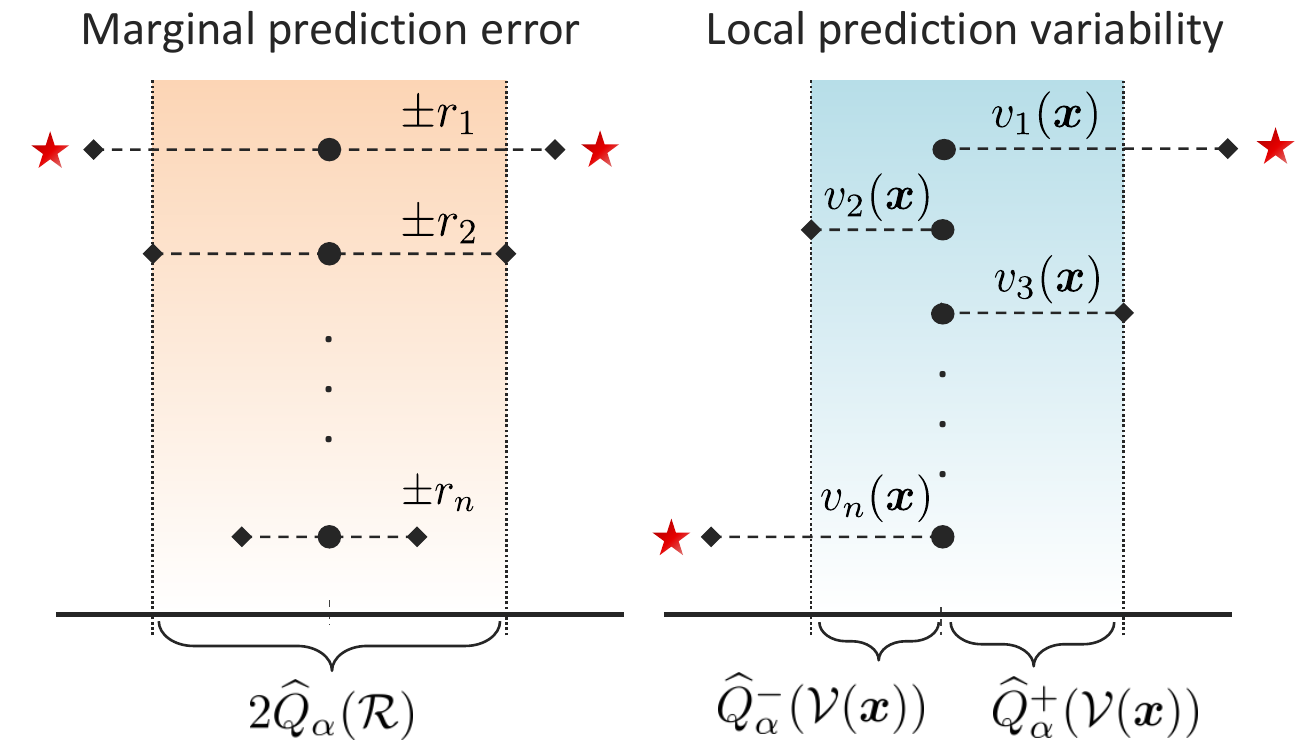}
  \caption{\footnotesize {\bf Illustration of the discriminative jackknife.} Confidence intervals are constructed using the empirical quantiles of the LOO residuals (left) and input-dependent prediction variability (right). Here, we depict the (sorted) elements of $\mathcal{R}$ and $\mathcal{V}$ --- elements marked with stars designate the boundaries of the $(1-\alpha)$ quantiles used to compute the DJ confidence intervals in (\ref{eq7}).} 
  \rule{\linewidth}{.75pt}    
  \label{Fig31} 
  \vspace{-8mm} 
\end{figure}

We construct a generic ameliorated jackknife,~the~DJ, which addresses the shortcomings of na\"ive~jackknife.~We first define some notation. Let the set $\mathcal{V}(\boldsymbol{x})$ be defined as:   
\begin{align}
\mathcal{V}(\boldsymbol{x}) = \left\{\, v_i(\boldsymbol{x}) \,\,|\,\, \forall i,\, 1 \leq i \leq n\,\right\},
\label{eq6}
\end{align}
where $v_i(\boldsymbol{x}) = f(\boldsymbol{x};\hat{\theta}) - f(\boldsymbol{x};\hat{\theta}_{-i})$.~Our~DJ~procedure~estimates the predictive confidence interval for a given test point $\boldsymbol{x}$ through the following steps:
\begin{align}
\widehat{\mathcal{C}}^{DJ}_{\alpha}(\boldsymbol{x};\hat{\theta}) &= [\,f_{-}(\boldsymbol{x};\hat{\theta}),\, f_{+}(\boldsymbol{x};\hat{\theta})\,], \nonumber \\
f_\gamma(\boldsymbol{x};\hat{\theta}) &= \mathcal{G}_{\alpha, \gamma}(\mathcal{R}, \mathcal{V}(\boldsymbol{x})),\, \gamma \in \{-1, +1\},\nonumber 
\end{align}
\vspace{-.25in}
\begin{align}
\mathcal{R} \Rightarrow \mbox{\footnotesize \bf Marginal Error},\, &\mathcal{V}(\boldsymbol{x}) \Rightarrow \mbox{\footnotesize \bf Local Variability},
\label{eq7}
\end{align} 
where $\mathcal{G}_{\alpha, \gamma}$ is a quantile function applied on the elements of the sets  of {\it marginal prediction errors} $\mathcal{R}$ and {\it local prediction variability} $\mathcal{V}$. The marginal prediction error terms use the LOO residuals to estimate the model's generalization error, and the prediction variability term quantifies the extent to which each training data point impacts the value of the prediction at test point $\boldsymbol{x}$. The prediction error is constant, i.e., does not depend on $\boldsymbol{x}$, hence it only contributes to coverage but does not contribute to discrimination. On the contrary, the local variability~term~depends~on~$\boldsymbol{x}$,~hence it fully determines the discrimination~performance.~The~function $\mathcal{G}_{\alpha, \gamma}$ can be constructed in a variety of ways; here we follow the Jackknife+ construct in \citep{barber2019predictive} 
\begin{align}
\mathcal{G}_{\alpha, \gamma}(\mathcal{R}, \mathcal{V}(\boldsymbol{x})) = \widehat{Q}^\gamma_{\alpha}(\{f(\boldsymbol{x};\hat{\theta})-v_i(\boldsymbol{x}) + \gamma \cdot r_i\}_i)
\label{eq7xxxx}
\end{align}
Figure \ref{Fig31} illustrates the construction of the DJ confidence intervals in (\ref{eq7}). The confidence intervals are chosen so that the boundaries of the average error and local variability are exceeded by $\lceil (n+1)(1-\alpha) \rceil$ out of the $n$ LOO samples  --- these are marked with a star. For the average prediction error term, the width of the resulting boundary is the same for any test data point $\boldsymbol{x} \in \mathcal{X}$. For the local prediction variability term, the width of the boundary depends on~$\boldsymbol{x}$,~and~should~be wider for less confident predictions, for which the model is vulnerable to the deletion of individual training points.

For the confidence interval's construction in (\ref{eq7}), it follows that the DJ interval width can be bounded above by
\begin{align}
W(\widehat{\mathcal{C}}^{DJ}_{\alpha}(\boldsymbol{x};\hat{\theta})) \,\leq \,2\, \widehat{Q}_{\alpha}(\mathcal{R}) + \sum_{\gamma}|\widehat{Q}^\gamma_{\alpha}(\mathcal{V}(\boldsymbol{x}))|.
\label{eq7rep}
\end{align} 
The marginal error and local variability terms in (\ref{eq7rep}) jointly capture two types of uncertainty: {\it epistemic} and {\it aleatoric} uncertainties \cite{gal2016uncertainty}. Epistemic uncertainty measures how well the model fits the data, and is reducible as the size of training data $n$ increases. On the contrary, aleatoric uncertainty is the irreducible variance arising from the inherent sources of ambiguity in the data, such as label noise or hidden features \cite{malinin2018predictive}. Consistency of the DJ confidence estimates requires that $W(\widehat{\mathcal{C}}^{DJ}_{\alpha}(\boldsymbol{x};\hat{\theta})) \to 0$, i.e., the interval width vanishes, as the size of the training data increases ($n \to \infty$). It follows from (\ref{eq7rep}) that if there is no aleatoric uncertainty, and if the underlying model is stable (i.e., $\lim_{n\to\infty} v_i = 0$) and consistent (i.e., $\lim_{n\to\infty} r_i = 0$), then the interval width $W(\widehat{\mathcal{C}}^{DJ}_{\alpha}(\boldsymbol{x};\hat{\theta}))$ vanishes as $n$ grows asymptotically (more training data is collected). 

\subsection{Efficient Implementation via Influence Functions}
\label{Sec32}
Exact computation of the DJ confidence intervals via (\ref{eq7}) requires re-training the model $n$~times~in order to collect the ``perturbed'' LOO parameters $\{\hat{\theta}_{-i}\}^{n}_{i=1}$. This exhaustive procedure is infeasible for large datasets and complex models. To scale up the~DJ,~we use {\it influence functions} --- a classic tool from robust statistics \cite{huber1981robust,hampel2011robust} --- in order to recover the parameters $\{\hat{\theta}_{-i}\}^{n}_{i=1}$ on the basis of the trained model $f(\boldsymbol{x};\hat{\theta})$, without the need for explicit re-training. Through this implementation, the DJ can be applied in a {\it post-hoc} fashion, requiring only knowledge of the model loss gradients.

Influence functions enable efficient computation of the effect of a training data point $(\boldsymbol{x}_i, y_i)$ on~$\hat{\theta}$.~This~is~achieved~by evaluating the change in $\hat{\theta}$ if $(\boldsymbol{x}_i, y_i)$ was up-weighted\footnote{A detailed technical background on influence functions and its connection with the jackknife is provided in Appendix A.} by some small $\epsilon$, resulting in a new parameter 
\begin{align}
\hat{\theta}_{i,\epsilon} \triangleq \arg \min_{\theta \in \Theta} L(\mathcal{D}_n, \theta) + \epsilon\cdot \ell(y_i, f(\boldsymbol{x}_i;\theta)). \nonumber
\end{align}
The rate of change in $\hat{\theta}$ due to an infinitesimal perturbation $\epsilon$ in data point $i$ is give by the (first-order) influence function 
\begin{equation}
\mathcal{I}^{(1)}_\theta(\boldsymbol{x}_i, y_i) = \frac{\partial\, \hat{\theta}_{i,\epsilon}}{\partial\, \epsilon}\, \, \Big|_{\epsilon = 0}.   
\label{eq8}
\end{equation}
Note that the model parameter $\hat{\theta}$ is a {\it statistical functional} of the data distribution $\mathbb{P}$. Perturbing the $i$-th training point is equivalent to perturbing $\mathbb{P}$ to create a new distribution $\mathbb{P}_{i,\epsilon} = (1 - \epsilon)\,\mathbb{P} + \epsilon \Delta (\boldsymbol{x}_i, y_i)$, where $\Delta (\boldsymbol{x}_i, y_i)$ denotes the Dirac distribution in the point $(\boldsymbol{x}_i, y_i)$. In this sense, the influence function in (\ref{eq8}) operationalizes the concept of derivatives to statistical functionals, i.e., the derivative of the parameters $\hat{\theta}$ with respect to the data distribution $\mathbb{P}$. 

By recognizing that influence functions are the ``derivatives'' of $\hat{\theta}$ with respect to $\mathbb{P}$, we can use a Taylor-type expansion to represent the counter-factual model parameter $\hat{\theta}_{i, \epsilon}$ (that would have been learned from a dataset with the $i$-th data point up-weighted) in terms of the parameter $\hat{\theta}$~(learned from the complete $\mathcal{D}_n$) as follows \cite{robins2008higher}:
\begin{equation}
\hat{\theta}_{i,\epsilon} = \hat{\theta} + \epsilon \cdot \mathcal{I}^{(1)}_\theta(\boldsymbol{x}_i, y_i) + \frac{\epsilon^2}{2!} \cdot \mathcal{I}^{(2)}_\theta(\boldsymbol{x}_i, y_i) +  \,\ldots    
\label{eq9}
\end{equation}
where $\mathcal{I}^{(k)}_\theta(\boldsymbol{x}_i, y_i)$ is the $k$-th order influence function, defined as $\mathcal{I}^{(k)}_\theta(\boldsymbol{x}_i, y_i) = \partial^k\, \hat{\theta}_{i,\epsilon} / \partial\, \epsilon^k \,|_{\epsilon = 0}$. The expansion in (\ref{eq9}), known as the {\it von Mises} expansion \cite{fernholz2012mises}, is a distributional analog of the Taylor expansion for statistical functionals. If all of the higher-order influence~functions~(HOIFs) in (\ref{eq9}) exist, then we can recover $\hat{\theta}_{i,\epsilon}$ without re-training the model on the perturbed training dataset. Since exact reconstruction of $\hat{\theta}_{i,\epsilon}$ requires an infinite number of HOIFs, we can only approximate $\hat{\theta}_{i,\epsilon}$ by including a finite number of HOIF terms from the von Mises expansion. %We discuss the impact of this approximation on the coverage and discrimination performance of the DJ in Section \ref{Sec34}. In Section \ref{Sec33}, we will derive a recursive formula for efficiently computing the HOIFs in (\ref{eq9}).

The LOO model parameters $\{\hat{\theta}_{-i}\}^n_{i=1}$, required for the construction of the DJ confidence~intervals,~can~be~obtained~by setting $\epsilon = -1/n$, i.e., $\hat{\theta}_{-i} = \hat{\theta}_{i,\frac{-1}{n}}$, since removing a training point is equivalent to up-weighting it by~$-1/n$~in~the loss function $L(\mathcal{D}_n;\theta)$. Thus, by setting $\epsilon = -1/n$ and selecting a prespecified number of HOIF terms $m$ for obtaining the approximate LOO parameters $\hat{\theta}^{(m)}_{-i}$, the DJ confidence intervals can be computed using the steps in Algorithm 1.

\subsection{Computing Influence Functions}
\label{Sec33}
The recent work on model interpretability in \cite{koh2017understanding} has studied the usage of influence functions to quantify the impact of individual data points on model training. There, first-order influence was computed, using a classical result in \cite{cook1982residuals}, as follows: 
\begin{equation}
\mathcal{I}^{(1)}_{\theta}(\boldsymbol{x}, y) = - H_{\theta}^{-1} \cdot \nabla_{\theta}\, \ell(y, f(\boldsymbol{x}, \theta)),  
\label{eq10}
\end{equation}
where $H_{\theta} \triangleq \nabla^2_{\theta}\, \sum_i\ell(y_i, f(\boldsymbol{x}_i, \theta))$  is the Hessian of the loss function, which is assumed to be positive definite. We derive an approximate expression for the second order influence function in terms of the first order influence as follows: 
\vspace{-0.1in}
\begin{equation}
\mathcal{I}^{(2)}_{\theta}(\boldsymbol{x}, y) \approx -2 H_{\theta}^{-1} \cdot \mathcal{I}^{(1)}_{\theta}(\boldsymbol{x}, y) \cdot \nabla^2_{\theta}\, \ell(y, f(\boldsymbol{x}, \theta)).  
\label{eq10xxx}
\end{equation}
In general, it can be shown that HOIFs can be recursively represented in terms of lower-order influence and loss gradients as \cite{giordano2019higher, debruyne2008model} %In the following Theorem\footnote{All proofs are provided in the supplementary appendix.}, we derive a recursive formula for computing HOIFs of the model parameters $\theta$.
\begin{align}
\mathcal{I}^{(k+1)}_{\theta} = - H_{\theta}^{-1}\Big(\sum^k_{m=1}g_m\big(\{\mathcal{I}^{(j)}_{\theta}, \nabla^{j}_{\theta}\, \ell_\theta\}^m_{j=1}\big)\Big),
\label{eq11new}
\end{align}
for some functions $\{g_m\}_m$. Here, we used short-hand notation for influence functions and loss gradients, dropping the dependency on $(\boldsymbol{x}, y)$. HOIFs exist if $\ell(.)$ is differentiable and locally convex in the neighborhood of $\theta$, and $H_{\theta} \succeq 0$.  In practice, we found that the second order terms are sufficient for obtaining an accurate estimate of the re-trained model parameters. The derivation of the second-order influence function in (\ref{eq10xxx}) is provided in Appendix B.

{\bf Computing HOIFs.}~On~the~positive~side,~(\ref{eq11new}) shows that we need to compute the inverse Hessian $H_{\theta}^{-1}$ only once for all HOIFs. However,~for~a~model~with~$p$~parameters, this is still a bottleneck operation with $\mathcal{O}(p^3)$ complexity. To address this hurdle, we capitalize on the recursive structure of (\ref{eq11new}) and the hessian-vector products approach in \cite{pearlmutter1994fast} to efficiently compute HOIFs as follows. To evaluate the $(k+1)$-th order influence given our estimate of $k$-th influence $\widetilde{\mathcal{I}}_k$, we execute the following steps: 

{\it (Step 1)} Compute the $k$-th order loss gradient $\nabla^k_\theta \ell_\theta$.

{\it (Step 2)} Evaluate $w = \sum^k_{m=1}g_m\big(\{\widetilde{\mathcal{I}}^{(j)}_{\theta}, \nabla^{j}_{\theta}\, \ell_\theta\}^m_{j=1}\big)$.

{\it (Step 3)} Sample $t$ data points $\{(\boldsymbol{x}_{s_i}, y_{s_i})\}_{i=1}^t$ from $\mathcal{D}_n$. 

{\it (Step 4)} Initialize $\widetilde{H}^{-1}_{0,\theta}w=w$, and recursively compute:
\begin{align}
\widetilde{H}^{-1}_{j,\theta}w = w + (\mathbf{I} - \nabla^2_\theta \ell_\theta)\cdot\widetilde{H}^{-1}_{j-1,\theta}w, \nonumber
\end{align} 
for $j \in \{0,\ldots,t\}$, where $\widetilde{H}^{-1}_{j,\theta} \triangleq \sum_{i=o}^j (\mathbf{I}-\widetilde{H}_{\theta})^i$, and $\widetilde{H}_{\theta}$ is the stochastic estimate of the Hessian computed over the sampled $t$ data points in $\{(\boldsymbol{x}_{s_i}, y_{s_i})\}_{i=1}^t$.

{\it (Step 5)} Return $\widetilde{\mathcal{I}}_{k+1} = \widetilde{\mathcal{I}}_{k} - \widetilde{H}^{-1}_{t,\theta}w$.

As shown through the steps above, the recursive nature of HOIFs allow us to reuse much of the computations involved in evaluating lower-order influence in computing higher-order terms. The stochastic estimation~process~above~is~motivated by the power series expansion of matrix inversion, and converges if $H_{\theta} \succeq 1$, which can always be ensured via appropriate scaling of the loss. We approximate the~higher~order loss gradients in Step 1 using coordinate-wise gradients, thus, for computing $m$ HOIFs, the overall complexity of the procedure above is linear in $n$, $m$ and $p$, i.e., $\mathcal{O}(npm)$.

\begin{algorithm}[t]
   \caption{The Discriminative Jackknife}
   \label{alg:example}
\begin{algorithmic}[1] 
   \STATE {\bfseries Input:} Learned parameter \mbox{\small $\hat{\theta}$}, influence order \mbox{\small $m$},
   \STATE $\,\,\,\,\,\,\,\,\,\,\,\,\,\,\,\,\,\,$ coverage \mbox{\small $\alpha$}, training data \mbox{\small $\mathcal{D}_n$}, test point \mbox{\small $\boldsymbol{x}$}.
   \vspace{0.025in} 
   \STATE {\bfseries Output:} DJ confidence interval \mbox{\small $\widehat{\mathcal{C}}^{DJ}_{\alpha}(\boldsymbol{x};\hat{\theta}, m)$}.\\ 
   \vspace{-0.05in}   
   \rule{8cm}{0.4pt}
   %\REPEAT
   %\STATE \mbox{\small \texttt{// Resampling sequence and interval blocks}}
   %\vspace{0.015in}
   \FOR{\mbox{\small $i=1$} {\bfseries to} \mbox{\small $n$}}
   \vspace{0.035in}
   \STATE \mbox{\small $\hat{\theta}^{\mbox{\tiny $(m)$}}_{-i} \gets \hat{\theta} - \sum^m_{k=1} (n^{-k} / k!)\, \cdot\, \mathcal{I}^{(k)}_{\theta}(\boldsymbol{x}_i, y_i)$.}
   \vspace{0.035in}
   \STATE \mbox{\small $r_{i} \gets \big|\,y_i - f(\boldsymbol{x}_i;\hat{\theta}^{\mbox{\tiny $(m)$}}_{-i})\,\big|$.}
   \vspace{0.035in}
   \STATE \mbox{\small $v_{i}(\boldsymbol{x}) \gets f(\boldsymbol{x};\hat{\theta}) - f(\boldsymbol{x};\hat{\theta}^{\mbox{\tiny $(m)$}}_{-i})$.}
   \vspace{0.035in}
   \ENDFOR
   \vspace{0.035in}
   \STATE \mbox{\small $f_{-}(\boldsymbol{x};\hat{\theta}) \gets \widehat{Q}^-_{\alpha}(\{f(\boldsymbol{x};\hat{\theta})-v_i(\boldsymbol{x}) + \gamma \cdot r_i\}_i)$.}
   \vspace{0.035in}
   \STATE \mbox{\small $f_{+}(\boldsymbol{x};\hat{\theta}) \gets \widehat{Q}^+_{\alpha}(\{f(\boldsymbol{x};\hat{\theta})-v_i(\boldsymbol{x}) + \gamma \cdot r_i\}_i)$.}
   \vspace{0.04in}   
   \STATE {\bf Return}\,\, \mbox{\small $\widehat{\mathcal{C}}^{DJ}_{\alpha, n}(\boldsymbol{x};\hat{\theta}, m) \gets [\,f_{-}(\boldsymbol{x};\hat{\theta}), f_{+}(\boldsymbol{x};\hat{\theta})\,]$.}
   \vspace{0.025in}
\end{algorithmic} 
\end{algorithm}

Despite the reduction in computational complexity, the proposed approximate procedure still entails a linear complexity in both the number of training data points $n$~and~the~number of model parameters $p$. This computational bottleneck limits our post-hoc procedure to relatively small networks, hence we regard our method's inability to {\it scale}~as its key limitation. Devising efficient methods for approximating~the Hessian is an interesting direction for future research.

\subsection{Theoretical Guarantees}
\label{Sec34}
We conclude this Section by revisiting the design requirements in Section \ref{Sec2}. In the following~Theorem,~we~show~that the DJ provides a guarantee on the coverage condition. 

{\bf Theorem 1.} {\it For any model $f(\boldsymbol{x};\hat{\theta})$, the coverage probability achieved by the approximate DJ with $m \to \infty$ is} 
\[
\mathbb{P}\left\{y \in \widehat{\mathcal{C}}^{DJ}_{\alpha}(\boldsymbol{x};\hat{\theta},\infty)\right\} \, \geq \, (1 - \,2\alpha).\,\,\,\,\,  \Box 
\]
Theorem 1 provides a strong, model-independent guarantee on the frequentist coverage of the DJ confidence intervals. In Section \ref{Sec5}, we show through empirical evaluation that in practice --- even with second order influence terms only --- the DJ intervals will achieve the target $(1-\alpha)$ coverage. With further assumptions on the algorithmic stability of the underlying model, it can also be shown that the exact DJ satisfies the discrimination condition in Section \ref{Sec22} (See Appendix C of the supplementary material). %This renders the DJ applicable to a wide range of deep learning models, with minimal assumptions needed on their algorithmic stability.  

\newpage
\section{Related Work}
\label{Sec4} 
Post-hoc methods for uncertainty quantification have been traditionally underexplored since existing approaches, such as calibration via temperature scaling \cite{platt1999probabilistic}, were known to under-perform compared to built-in methods \cite{ovadia2019can}. However, recent works have revived post-hoc approaches using ideas based on bootstrapping \cite{schulam2019can}, jackknife resampling \cite{barber2019predictive,giordano2018swiss} and cross-validation \cite{vovk2018cross,barber2019conformal}. An overview of the different classes of post-hoc and built-in methods proposed in recent literature is provided in Table \ref{Table41}.
\begin{table}[h]
\centering
\vspace{-4mm}
\caption{{\footnotesize Overview of existing uncertainty quantification methods.}}
{\footnotesize
\begin{tabular}{@{}lcccc@{}}
\toprule
\textbf{Method} & \textbf{\begin{tabular}[c]{@{}l@{}}Bayesian /\\ Frequentist\end{tabular}} & \textbf{\begin{tabular}[c]{@{}l@{}}Post-hoc / \\Built-in\end{tabular}} & \textbf{\begin{tabular}[c]{@{}l@{}}Coverage \end{tabular}} \\ \midrule
Bayesian neural nets & Bayesian &  Built-in  &  None \\ 
Prob. backprop. & Bayesian &  Built-in  &  None \\ 
Monte Carlo dropout & Bayesian &  Built-in  &  None \\
Deep ensembles & Frequentist &  Built-in  &  None \\
RUE & Frequentist &  Post-hoc  &  None \\ \midrule
DJ & Frequentist &  Post-hoc  & $1-2\alpha$ \\
\bottomrule
\end{tabular}}
\vspace{-4mm}
\label{Table41}
\end{table}

Bayesianism is the dominant approach to uncertainty quantification in deep learning \cite{welling2011bayesian,hernandez2015probabilistic, ritter2018scalable, maddox2019simple}. A post-hoc application of Bayesian methods is not possible since by their very nature, Bayesian models require specifying priors over model parameters, which leads to major modifications in the inference algorithms. While Bayesian models provide a formal framework for uncertainty estimation, posterior credible intervals do not guarantee frequentist coverage, and more crucially, the achieved coverage can be very sensitive to hyper-parameter tuning \cite{bayarri2004interplay}. Moreover, exact Bayesian inference is computationally prohibitive, and alternative approximations --- e.g., \cite{gal2016dropout} --- may induced posterior distributions that do not concentrate asymptotically \cite{osband2016risk, hron2017variational}. 

Deep ensembles \cite{lakshminarayanan2017simple} are regarded as the most competitive (non-Bayesian) benchmark for uncertainty~estimation~\cite{ovadia2019can}.~This~method repeatedly re-trains the model on sub-samples of the data (using adversarial training), and then estimates uncertainty through the variance of the aggregate predictions. A similar bootstrapping approach developed in \cite{schulam2019can}, dubbed resampling uncertainty estimation (RUE), uses the model's Hessian and loss gradients to create an ensemble without re-training.~While~these~methods~may~perform favorably in terms of discrimination, they are likely to undercover, since they only consider local variability terms akin to those in (\ref{eq7}). Additionally, ensemble methods do not provide theoretical guarantees on their performance.

\begin{figure*}[t]
  \centering
  \includegraphics[width=6.75in]{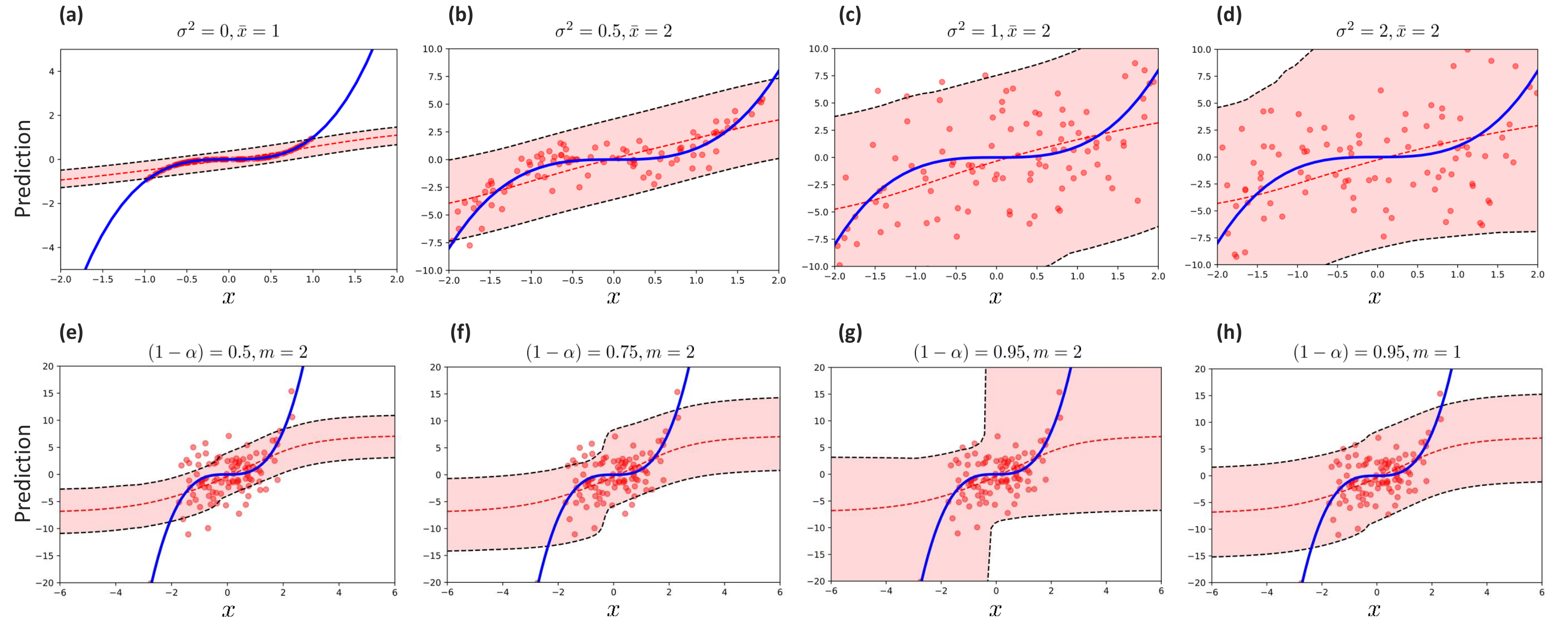}
  \caption{\footnotesize {\bf DJ confidence intervals in a one-dimensional feature space.} {\bf (a)} Uniform feature distribution with $\bar{x}=1$ and no aleatoric noise. {\bf (b)} For a Uniform feature distribution with $\bar{x}=2$ and noise variance $\sigma^2=0.5$, the DJ confidence intervals are wider than those in (a). The confidence intervals are of a fixed width for all $x$ because the training points are uniformly distributed everywhere in the feature space. {\bf (c)} For a Uniform feature distribution with $\bar{x}=2$ and noise variance $\sigma^2=1$, the DJ confidence intervals are wider than (b) and some of the training data points are not covered because of the high noise variance. {\bf (d)} Uniform feature distribution with $\bar{x}=2$ and noise variance $\sigma^2=2$. {\bf (e)} Normal feature distribution with noise variance $\sigma^2=1$ and target coverage $(1-\alpha)=0.5$. {\bf (f)} Normal feature distribution with noise variance $\sigma^2=1$ and target coverage $(1-\alpha)=0.75$. Because of the stricter coverage target, the DJ confidence intervals are wider than those in (e). {\bf (g)} Normal feature distribution with noise variance $\sigma^2=1$ and target coverage $(1-\alpha)=0.95$. Because the normal feature distribution is associated with epistemic uncertainty, the width of the confidence interval is not uniform for the different values of $x$. {\bf (h)} DJ confidence intervals with first-order influence functions ($m=1$) for the same setting in (g). Here, we can see that the DJ confidence intervals exhibit a fixed width for all values of the feature $x$, and do not capture epistemic uncertainty as in (g).} 
  \rule{\linewidth}{.75pt}   
  \label{Fig_new_1} 
  \vspace{-7mm} 
\end{figure*} 

The (infinitesimal) jackknife method was previously used for quantifying the predictive uncertainty in random forests \cite{wager2014confidence,mentch2016quantifying,wager2018estimation}. In these works, however, the developed jackknife estimators are bespoke to bagging predictors, and cannot be straightforwardly extended to deep neural networks. More recently, general-purpose jackknife estimators were developed in \cite{barber2019predictive}, where two exhaustive leave-one-out procedures: the {\it jackknife+} and the {\it jackknife-minmax} where shown to have assumption-free worst-case coverage guarantees of $(1-2\alpha)$ and $(1-\alpha)$, respectively. Our work improves on these results by alleviating the need for exhaustive leave-one-out re-training. 

\section{Experiments}
\label{Sec5}
In this Section, we evaluate the DJ using synthetic and real data, and compare its performance with various baselines. Further experimental details are deferred to Appendix D.

{\bf Baselines.} We compared our DJ~method~with~4~state~of~the art baselines. These included 3 built-in Bayesian methods: (1) Monte Carlo Dropout (MCDP) \cite{gal2016dropout}, (2) Probabilistic backpropagation (PBP) \cite{hernandez2015probabilistic}, and (3) Bayesian neural networks with inference via stochastic gradient Langevin dynamics (BNN-SGLD) \cite{welling2011bayesian}. In addition, we considered deep ensembles (DE) \cite{lakshminarayanan2017simple}, which is deemed the most competitive built-in frequentist method \cite{ovadia2019can}. For a target coverage of $(1-\alpha)$, uncertainty estimates were obtained by setting posterior quantile functions to $(1-\alpha)$ (for Bayesian methods), or obtaining the $(1-\alpha)$ percentile point of a normal distribution (for frequentist methods).~Details~on~the implementation, hyper-parameter tuning and uncertainty calibration of all baselines are provided in Appendix D. 

{\bf Evaluation metrics.} In all experiments, we used the mean squared error (MSE) as the loss~$L(\mathcal{D}_n, \hat{\theta})$~for~training~the model $f(\boldsymbol{x};\hat{\theta})$.~To~ensure~fair~comparisons,~the~hyperparameters of the underlying neural network $f(\boldsymbol{x};\hat{\theta})$~were~fixed for all baselines. In each experiment, the uncertainty estimate $\widehat{\mathcal{C}}_{\alpha}(\boldsymbol{x};\hat{\theta})$ is obtained from a~training~sample,~and~then coverage and discrimination are evaluated on a test sample. To evaluate empirical coverage probability, we compute the fraction of test samples for which $y$ resides in $\widehat{\mathcal{C}}_{\alpha}(\boldsymbol{x}; \hat{\theta})$. Discrimination was evaluated as follows. For~each~baseline~method, we evaluate the interval~width~$W(\widehat{\mathcal{C}}_{\alpha}(\boldsymbol{x};\hat{\theta}))$~for all test points. For a given error threshold $\mathcal{E}$, we use the interval width to detect whether the test prediction~$f(\boldsymbol{x}; \hat{\theta})$~is a high-confidence, i.e., $\ell(y, f(\boldsymbol{x}; \theta)) \leq \mathcal{E}$, or low-confidence prediction, i.e., $\ell(y, f(\boldsymbol{x}; \theta)) > \mathcal{E}$.~We~use~the~area~under~the precision-recall curve (AUPRC) --- also known as the average precision score --- in order to evaluate the accuracy of this confidence classification task.

\subsection{Synthetic Data} 
\label{Sec61}
We start off by illustrating the DJ confidence intervals using data generated from the following synthetic model. In particular, we use the synthetic model introduced in \cite{hernandez2015probabilistic}, defined as follows:
\begin{align}
y = x^3 + \epsilon,
\end{align}
where $\epsilon \sim \mathcal{N}(0, \sigma^2)$, and $\sigma^2$ is the noise variance.~We~consider two possible feature distributions:
\begin{align}
\mbox{Uniform feature distribution}:\,\,& x \sim \mbox{U}([-\bar{x}, \bar{x}]), \nonumber \\ 
\mbox{Normal feature distribution}:\,\,& x \sim \mathcal{N}(0, \bar{\sigma}^2_x). 
\label{eqSec61new}
\end{align}
Under the uniform distribution, the model will be equally uncertain about its predictions for any $x$ since all feature instances have the same amount of noise and frequency of observations. In this case, the main source of uncertainty is the aleatoric uncertainty resulting from the noise $\epsilon$. On the contrary, under the normal distribution, the model will be more uncertain in predictions made for values of $x$ that deviate from 0. This is because most of the~training~data~will be concentrated around 0, leading to an increased epistemic uncertainty for very large or very small values of $x$. In all experiments, we fit a 2-layer feed-forward neural network with 100 hidden units and compute the DJ confidence intervals using the post-hoc procedure in Algorithm 1. 

\begin{figure*}[t]
\centering
  \includegraphics[width=6.75in]{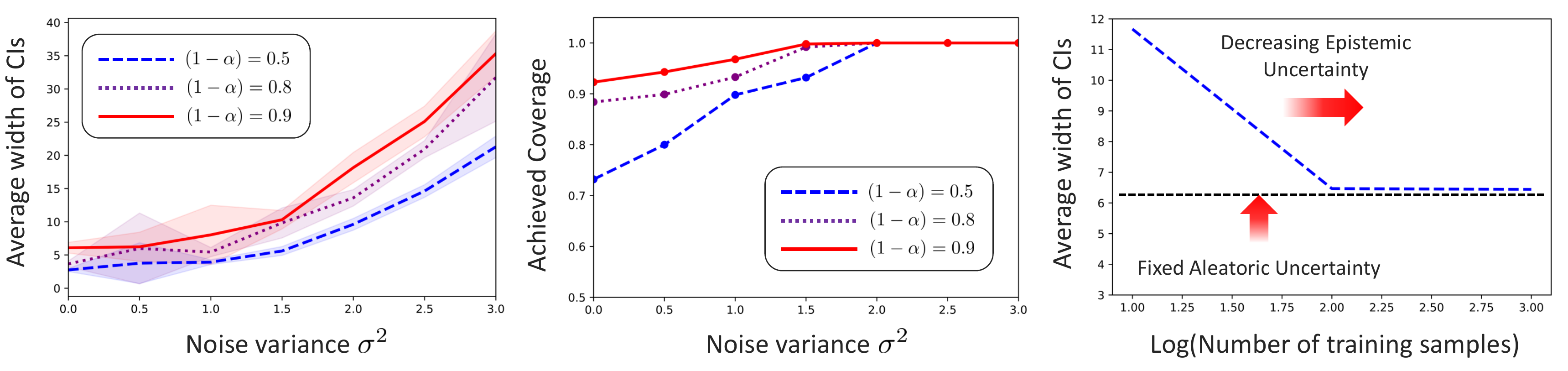}
  \caption{{\footnotesize Average width of the DJ confidence intervals and achieved coverage at varying levels of aleatoric and epistemic uncertainty.}}   
  \label{Fig52} 
  \rule{\linewidth}{.75pt} 
  \vspace{-8mm}
\end{figure*} 

{\bf Results.} In Figure \ref{Fig_new_1}, we depict various samples of the DJ confidence intervals for different feature distributions, target coverage levels, and noise variances. In~Figures~\ref{Fig_new_1}~(a)~to~(d), we show the confidence intervals issued by a model trained under the uniform feature distribution: here, we can see that the interval width does not vary significantly for the different values of $x$, because the training points are uniformly distributed everywhere in the feature space. Moreover, the interval width increases as the noise variance increases.

In Figures~\ref{Fig_new_1}~(e)~to~(h),~we~show~the~DJ~confidence~intervals issued by a model trained under the normal feature distribution. Here, we see that the interval width is narrowest around $x=0$, i.e., the point around which most training points are concentrated. We also see that the inclusion of the second-order influence terms enriches the shape of the confidence intervals in a way that reflects the ground-truth epistemic uncertainty (see Figures \ref{Fig_new_1} (g) and (h)). 

Finally, in Figure \ref{Fig52}, we show the average width of the DJ confidence intervals (averaged over 100 test points across 10 simulations) and the achieved coverage for a neural network model trained using $n=100$ training points with varying levels of ground-truth aleatoric uncertainty (i.e., varying noise variance $\sigma^2$). As we can see, the width of the confidence intervals increase for larger noise variances (reflecting higher levels of reported uncertainty) and for more strict target coverage $(1-\alpha)$ (Figure \ref{Fig52}, left). For all noise variances, the DJ confidence intervals manage to achieve the target coverage levels (Figure \ref{Fig52}, middle). By changing the number of training points with a fixed noise variance, we control for the amount of epistemic uncertainty: as we can see in Figure \ref{Fig52} (right), the width of the confidence intervals decrease as the training data increases, until it hits a floor dictated by the inherent aleatoric uncertainty in the labels. 

\subsection{Real Data: Auditing Model Reliability}
\label{Sec62}
In this Section, we conduct a series of experiments on real-world datasets in order to evaluate the accuracy of uncertainty estimates issued by the DJ procedure.~In~particular,~we show how uncertainty estimates can be used to audit the reliability of a model using experiments on 4 UCI benchmark datasets for regression: yacht hydrodynamics (Yacht), Boston housing (Housing), energy efficiency (Energy) and naval propulsion (Naval) \cite{Dua:2019}.  

In each experiment, we use 80$\%$ of the data for training and 20$\%$ for testing. We use a 2-layer~neural~network~model~with 100 hidden units, Tanh activation functions,~MSE~loss,~and~a single set of learning hyper-parameters for all baselines (1000 epochs with 100 samples per minibatch, and an Adam optimizer with default settings). We set the target coverage to $(1-\alpha)=0.9$. On each test set, we evaluate the model's MSE, achieved coverage rate and AUPRC in predicting whether the model's test error exceeds a threshold $\mathcal{E}$ that is set to be 90$\%$ percentile of the empirical distribution over test errors. Results are provided in Table \ref{Table51}. 

We observe that, by virtue of its post-hoc~nature,~the~DJ~procedure yields the best predictive performance~(MSE)~on almost all baselines. This is because the DJ does not interfere with the model training or compromise its accuracy, and is only applied on an already trained model that is optimized to minimize the MSE. The post-hoc nature of our method does not compromise the accuracy of its uncertainty intervals. Across all data sets, the DJ achieves the desired target coverage, whereas other Bayesian methods (e.g., BNN and PBP) tend to under-cover the true labels. Moreover, DJ provides high AUPRC scores on all data sets, and on data sets were its AUPRC scores are lower than the other coverage-achieving baselines (e.g., MCDP), it offers a much better predictive accuracy in terms of the MSE.

%\begin{tcolorbox}[tab3,tabularx={lcc}]
\begin{table}[t]
\centering
{\small
\begin{tabular}{@{}lccccc@{}}
\toprule[1.25pt]
\textbf{}         & \multicolumn{4}{c}{\textbf{Dataset}}                                                                                                                                                                                                                              \\ \midrule
\textbf{Method}   & \multicolumn{1}{c}{\textbf{Yacht}} & \multicolumn{1}{c}{\textbf{Housing}} & \multicolumn{1}{c}{\textbf{Energy}} & \multicolumn{1}{c}{\textbf{Kin8nm}} \\ \midrule[1.25pt]
\textbf{DJ} & {\scriptsize 0.87 $\pm$ 0.05}  & {\scriptsize 0.80 $\pm$ 0.04}  & {\scriptsize 0.77 $\pm$ 0.08}  & {\scriptsize 0.88 $\pm$ 0.01}   \\
 & ({\scriptsize 95.9$\%$})$^*$  &  ({\scriptsize 99.8$\%$})$^*$  &  ({\scriptsize 98.11$\%$})$^*$ & ({\scriptsize 93.77$\%$})$^*$  \\ 
  & {\scriptsize 26.55}  &  {\scriptsize 33.87}  & {\scriptsize 11.06} & {\scriptsize 0.00}  \\ \midrule
\textbf{MCDP} &   {\scriptsize 0.67 $\pm$ 0.06}  & {\scriptsize 0.86 $\pm$ 0.00}                                       &  {\scriptsize 0.84 $\pm$ 0.03} & {\scriptsize 0.83 $\pm$ 0.03}  \\
      & ({\scriptsize 100.0$\%$})$^*$ & ({\scriptsize 99.6$\%$})$^*$  & ({\scriptsize 100.0$\%$})$^*$ & ({\scriptsize 100.0$\%$}) \\ 
        & {\scriptsize 150.93}  &  {\scriptsize 113.04} &  {\scriptsize 92.57} & {\scriptsize 0.05}  \\ \midrule     
\textbf{PBP} &   {\scriptsize 0.66 $\pm$ 0.06}  & {\scriptsize 0.85 $\pm$ 0.03}                                       &  {\scriptsize 0.84 $\pm$ 0.04} & {\scriptsize 0.82 $\pm$ 0.04}  \\
      & ({\scriptsize 70.4$\%$}) & ({\scriptsize 5.0$\%$})  & ({\scriptsize 10.1$\%$}) & ({\scriptsize 89.9$\%$})  \\ 
        & {\scriptsize 22.21}  &  {\scriptsize 221.11}  & {\scriptsize 201.73} & {\scriptsize 0.62}   \\ \midrule 
\textbf{DE} &   {\scriptsize 0.87 $\pm$ 0.04}  & {\scriptsize 0.62 $\pm$ 0.04}                                       &  {\scriptsize 0.80 $\pm$ 0.09} & {\scriptsize 0.82 $\pm$ 0.02}      \\
      & ({\scriptsize 0.0$\%$}) & ({\scriptsize 19.4$\%$})  & ({\scriptsize 23.6$\%$}) & ({\scriptsize 21.83$\%$})  \\
        & {\scriptsize 327.74}  &  {\scriptsize 61.82}  &  {\scriptsize 21.53} & {\scriptsize 0.03}  \\ \midrule 
\textbf{BNN} & {\scriptsize 0.81 $\pm$ 0.05} & {\scriptsize 0.88 $\pm$ 0.00}                                       &  &  {\scriptsize 0.89 $\pm$ 0.00} \\        
      & ({\scriptsize 82.3$\%$}) & ({\scriptsize 89.0$\%$})  & --- & ({\scriptsize 100.0$\%$})   \\
        & {\scriptsize 317.94}  &  {\scriptsize 118.89}  &  & {\scriptsize 0.68}   \\ \bottomrule[1.25pt]
\end{tabular}}
\caption{{\bf Auditing predictive model reliability.} AUPRC performance ($\pm$ 95$\%$~confidence intervals) of all baselines on the real-world UCI regression datasets. In each cell, the first line contains the AUPRC score, the second line contains the achieved (empirical) coverage and the third line lists the MSE loss. Coverage rates marked with an asterisk achieve the desired 90$\%$ target rate. Blank entries correspond to models with confidence intervals that perform no better than random guessing with respect to the AUPRC.}
\vspace{-1mm}
\label{Table51}
  \rule{\linewidth}{.75pt} 
  \vspace{-8mm}
\end{table}

\section{Conclusion}
Uncertainty quantification is a crucial requirement in various high-stakes applications,~wherein~deep~learning~can~inform critical decision-making. In this paper, we introduced a rigorous frequentist procedure for quantifying the uncertainty in predictions issued by deep learning models in a post-hoc fashion. Our procedure, which is inspired by classical jackknife estimators, does not require any modifications in the underlying deep learning model, and provides theoretical guarantees on~its~achieved performance.~Because~of~its post-hoc and model-agnostic nature, this procedure can be applied to a wide variety of models ranging from feed-forward networks to convolutional and recurrent~networks.~While~our focus was mainly on deep learning models, our procedure can also be applied to general machine learning models. 

A key ingredient of our procedure is the usage of influence functions to reconstruct leave-one-out model parameters without the need for explicit re-training. Influence functions provide a powerful tool for constructing ensembles of models in a post-hoc fashion that can be used to assess model variability without the need for built-in designs for uncertainty intervals. While we present an algorithm that recursively computes influence functions with a complexity that is linear in the number of model parameters and size of training data, our procedure is still limited to relatively small networks or small data sets. Developing methods for fast computation of the Hessian matrix that would scale up our method to more complex networks and larger data sets is a valuable direction for future research.    

\section*{Acknowledgments}
The authors would like to thank the reviewers for their helpful comments. This work was supported by the US Office of Naval Research (ONR) and the National Science Foundation (NSF grants 1524417 and 1722516).

%\nocite{*}

\bibliography{DJackknife_arxiv}

\begin{thebibliography}{44}
\providecommand{\natexlab}[1]{#1}
\providecommand{\url}[1]{\texttt{#1}}
\expandafter\ifx\csname urlstyle\endcsname\relax
  \providecommand{\doi}[1]{doi: #1}\else
  \providecommand{\doi}{doi: \begingroup \urlstyle{rm}\Url}\fi

\bibitem[Alaa \& van~der Schaar(2017)Alaa and van~der Schaar]{alaa2017bayesian}
Alaa, A.~M. and van~der Schaar, M.
\newblock Bayesian inference of individualized treatment effects using
  multi-task gaussian processes.
\newblock In \emph{Advances in Neural Information Processing Systems}, pp.\
  3424--3432, 2017.

\bibitem[Amodei et~al.(2016)Amodei, Olah, Steinhardt, Christiano, Schulman, and
  Man{\'e}]{amodei2016concrete}
Amodei, D., Olah, C., Steinhardt, J., Christiano, P., Schulman, J., and
  Man{\'e}, D.
\newblock Concrete problems in ai safety.
\newblock \emph{arXiv preprint arXiv:1606.06565}, 2016.

\bibitem[Barber et~al.(2019{\natexlab{a}})Barber, Candes, Ramdas, and
  Tibshirani]{barber2019conformal}
Barber, R.~F., Candes, E.~J., Ramdas, A., and Tibshirani, R.~J.
\newblock Conformal prediction under covariate shift.
\newblock \emph{arXiv preprint arXiv:1904.06019}, 2019{\natexlab{a}}.

\bibitem[Barber et~al.(2019{\natexlab{b}})Barber, Candes, Ramdas, and
  Tibshirani]{barber2019predictive}
Barber, R.~F., Candes, E.~J., Ramdas, A., and Tibshirani, R.~J.
\newblock Predictive inference with the jackknife+.
\newblock \emph{arXiv preprint arXiv:1905.02928}, 2019{\natexlab{b}}.

\bibitem[Bayarri \& Berger(2004)Bayarri and Berger]{bayarri2004interplay}
Bayarri, M.~J. and Berger, J.~O.
\newblock The interplay of bayesian and frequentist analysis.
\newblock \emph{Statistical Science}, pp.\  58--80, 2004.

\bibitem[Bousquet \& Elisseeff(2002)Bousquet and
  Elisseeff]{bousquet2002stability}
Bousquet, O. and Elisseeff, A.
\newblock Stability and generalization.
\newblock \emph{Journal of machine learning research}, 2\penalty0
  (Mar):\penalty0 499--526, 2002.

\bibitem[Cohn et~al.(1996)Cohn, Ghahramani, and Jordan]{cohn1996active}
Cohn, D.~A., Ghahramani, Z., and Jordan, M.~I.
\newblock Active learning with statistical models.
\newblock \emph{Journal of artificial intelligence research}, 4:\penalty0
  129--145, 1996.

\bibitem[Cook \& Weisberg(1982)Cook and Weisberg]{cook1982residuals}
Cook, R.~D. and Weisberg, S.
\newblock \emph{Residuals and influence in regression}.
\newblock New York: Chapman and Hall, 1982.

\bibitem[Debruyne et~al.(2008)Debruyne, Hubert, and Suykens]{debruyne2008model}
Debruyne, M., Hubert, M., and Suykens, J.~A.
\newblock Model selection in kernel based regression using the influence
  function.
\newblock \emph{Journal of Machine Learning Research}, 9\penalty0
  (Oct):\penalty0 2377--2400, 2008.

\bibitem[Devroye \& Wagner(1979)Devroye and Wagner]{devroye1979distribution}
Devroye, L. and Wagner, T.
\newblock Distribution-free inequalities for the deleted and holdout error
  estimates.
\newblock \emph{IEEE Transactions on Information Theory}, 25\penalty0
  (2):\penalty0 202--207, 1979.

\bibitem[Dua \& Graff(2017)Dua and Graff]{Dua:2019}
Dua, D. and Graff, C.
\newblock {UCI} machine learning repository, 2017.
\newblock URL \url{http://archive.ics.uci.edu/ml}.

\bibitem[Dusenberry et~al.(2019)Dusenberry, Tran, Choi, Kemp, Nixon, Jerfel,
  Heller, and Dai]{dusenberry2019analyzing}
Dusenberry, M.~W., Tran, D., Choi, E., Kemp, J., Nixon, J., Jerfel, G., Heller,
  K., and Dai, A.~M.
\newblock Analyzing the role of model uncertainty for electronic health
  records.
\newblock \emph{arXiv preprint arXiv:1906.03842}, 2019.

\bibitem[Efron(1992)]{efron1992jackknife}
Efron, B.
\newblock Jackknife-after-bootstrap standard errors and influence functions.
\newblock \emph{Journal of the Royal Statistical Society: Series B
  (Methodological)}, 54\penalty0 (1):\penalty0 83--111, 1992.

\bibitem[Fernholz(2012)]{fernholz2012mises}
Fernholz, L.~T.
\newblock \emph{Von Mises calculus for statistical functionals}, volume~19.
\newblock Springer Science \& Business Media, 2012.

\bibitem[Gal(2016)]{gal2016uncertainty}
Gal, Y.
\newblock \emph{Uncertainty in deep learning}.
\newblock PhD thesis, PhD thesis, University of Cambridge, 2016.

\bibitem[Gal \& Ghahramani(2016)Gal and Ghahramani]{gal2016dropout}
Gal, Y. and Ghahramani, Z.
\newblock Dropout as a bayesian approximation: Representing model uncertainty
  in deep learning.
\newblock In \emph{International Conference on Machine Learning (ICML)}, pp.\
  1050--1059, 2016.

\bibitem[Giordano et~al.(2018)Giordano, Stephenson, Liu, Jordan, and
  Broderick]{giordano2018swiss}
Giordano, R., Stephenson, W., Liu, R., Jordan, M.~I., and Broderick, T.
\newblock A swiss army infinitesimal jackknife.
\newblock \emph{arXiv preprint arXiv:1806.00550}, 2018.

\bibitem[Giordano et~al.(2019)Giordano, Jordan, and
  Broderick]{giordano2019higher}
Giordano, R., Jordan, M.~I., and Broderick, T.
\newblock A higher-order swiss army infinitesimal jackknife.
\newblock \emph{arXiv preprint arXiv:1907.12116}, 2019.

\bibitem[Hampel et~al.(2011)Hampel, Ronchetti, Rousseeuw, and
  Stahel]{hampel2011robust}
Hampel, F.~R., Ronchetti, E.~M., Rousseeuw, P.~J., and Stahel, W.~A.
\newblock \emph{Robust statistics: the approach based on influence functions},
  volume 196.
\newblock John Wiley \& Sons, 2011.

\bibitem[Hern{\'a}ndez-Lobato \& Adams(2015)Hern{\'a}ndez-Lobato and
  Adams]{hernandez2015probabilistic}
Hern{\'a}ndez-Lobato, J.~M. and Adams, R.
\newblock Probabilistic backpropagation for scalable learning of bayesian
  neural networks.
\newblock In \emph{International Conference on Machine Learning (ICML)}, pp.\
  1861--1869, 2015.

\bibitem[Hron et~al.(2017)Hron, Matthews, and Ghahramani]{hron2017variational}
Hron, J., Matthews, A. G. d.~G., and Ghahramani, Z.
\newblock Variational gaussian dropout is not bayesian.
\newblock \emph{arXiv preprint arXiv:1711.02989}, 2017.

\bibitem[Huber \& Ronchetti(1981)Huber and Ronchetti]{huber1981robust}
Huber, P.~J. and Ronchetti, E.~M.
\newblock Robust statistics john wiley \& sons.
\newblock \emph{New York}, 1\penalty0 (1), 1981.

\bibitem[Kingma et~al.(2015)Kingma, Salimans, and
  Welling]{kingma2015variational}
Kingma, D.~P., Salimans, T., and Welling, M.
\newblock Variational dropout and the local reparameterization trick.
\newblock In \emph{Advances in Neural Information Processing Systems}, pp.\
  2575--2583, 2015.

\bibitem[Koh \& Liang(2017)Koh and Liang]{koh2017understanding}
Koh, P.~W. and Liang, P.
\newblock Understanding black-box predictions via influence functions.
\newblock In \emph{Proceedings of the 34th International Conference on Machine
  Learning-Volume 70}, pp.\  1885--1894. JMLR. org, 2017.

\bibitem[Lakshminarayanan et~al.(2017)Lakshminarayanan, Pritzel, and
  Blundell]{lakshminarayanan2017simple}
Lakshminarayanan, B., Pritzel, A., and Blundell, C.
\newblock Simple and scalable predictive uncertainty estimation using deep
  ensembles.
\newblock In \emph{Advances in Neural Information Processing Systems
  (NeurIPS)}, pp.\  6402--6413, 2017.

\bibitem[Lawless \& Fredette(2005)Lawless and Fredette]{lawless2005frequentist}
Lawless, J. and Fredette, M.
\newblock Frequentist prediction intervals and predictive distributions.
\newblock \emph{Biometrika}, 92\penalty0 (3):\penalty0 529--542, 2005.

\bibitem[LeCun et~al.(2015)LeCun, Bengio, and Hinton]{lecun2015deep}
LeCun, Y., Bengio, Y., and Hinton, G.
\newblock Deep learning.
\newblock \emph{nature}, 521\penalty0 (7553):\penalty0 436, 2015.

\bibitem[Leonard et~al.(1992)Leonard, Kramer, and Ungar]{leonard1992neural}
Leonard, J., Kramer, M.~A., and Ungar, L.
\newblock A neural network architecture that computes its own reliability.
\newblock \emph{Computers \& chemical engineering}, 16\penalty0 (9):\penalty0
  819--835, 1992.

\bibitem[Maddox et~al.(2019)Maddox, Garipov, Izmailov, Vetrov, and
  Wilson]{maddox2019simple}
Maddox, W., Garipov, T., Izmailov, P., Vetrov, D., and Wilson, A.~G.
\newblock A simple baseline for bayesian uncertainty in deep learning.
\newblock \emph{arXiv preprint arXiv:1902.02476}, 2019.

\bibitem[Malinin \& Gales(2018)Malinin and Gales]{malinin2018predictive}
Malinin, A. and Gales, M.
\newblock Predictive uncertainty estimation via prior networks.
\newblock In \emph{Advances in Neural Information Processing Systems}, pp.\
  7047--7058, 2018.

\bibitem[Mentch \& Hooker(2016)Mentch and Hooker]{mentch2016quantifying}
Mentch, L. and Hooker, G.
\newblock Quantifying uncertainty in random forests via confidence intervals
  and hypothesis tests.
\newblock \emph{The Journal of Machine Learning Research (JMLR)}, 17\penalty0
  (1):\penalty0 841--881, 2016.

\bibitem[Miller(1974)]{miller1974jackknife}
Miller, R.~G.
\newblock The jackknife-a review.
\newblock \emph{Biometrika}, 61\penalty0 (1):\penalty0 1--15, 1974.

\bibitem[Osband(2016)]{osband2016risk}
Osband, I.
\newblock Risk versus uncertainty in deep learning: Bayes, bootstrap and the
  dangers of dropout.
\newblock In \emph{NIPS Workshop on Bayesian Deep Learning}, 2016.

\bibitem[Ovadia et~al.(2019)Ovadia, Fertig, Ren, Nado, Sculley, Nowozin,
  Dillon, Lakshminarayanan, and Snoek]{ovadia2019can}
Ovadia, Y., Fertig, E., Ren, J., Nado, Z., Sculley, D., Nowozin, S., Dillon,
  J.~V., Lakshminarayanan, B., and Snoek, J.
\newblock Can you trust your model's uncertainty? evaluating predictive
  uncertainty under dataset shift.
\newblock \emph{arXiv preprint arXiv:1906.02530}, 2019.

\bibitem[Pearlmutter(1994)]{pearlmutter1994fast}
Pearlmutter, B.~A.
\newblock Fast exact multiplication by the hessian.
\newblock \emph{Neural computation}, 6\penalty0 (1):\penalty0 147--160, 1994.

\bibitem[Platt et~al.(1999)]{platt1999probabilistic}
Platt, J. et~al.
\newblock Probabilistic outputs for support vector machines and comparisons to
  regularized likelihood methods.
\newblock \emph{Advances in large margin classifiers}, 10\penalty0
  (3):\penalty0 61--74, 1999.

\bibitem[Ritter et~al.(2018)Ritter, Botev, and Barber]{ritter2018scalable}
Ritter, H., Botev, A., and Barber, D.
\newblock A scalable laplace approximation for neural networks.
\newblock 2018.

\bibitem[Robins et~al.(2008)Robins, Li, Tchetgen, van~der Vaart,
  et~al.]{robins2008higher}
Robins, J., Li, L., Tchetgen, E., van~der Vaart, A., et~al.
\newblock Higher order influence functions and minimax estimation of nonlinear
  functionals.
\newblock In \emph{Probability and statistics: essays in honor of David A.
  Freedman}, pp.\  335--421. Institute of Mathematical Statistics, 2008.

\bibitem[Schulam \& Saria(2019)Schulam and Saria]{schulam2019can}
Schulam, P. and Saria, S.
\newblock Can you trust this prediction? auditing pointwise reliability after
  learning.
\newblock In \emph{The 22nd International Conference on Artificial Intelligence
  and Statistics}, pp.\  1022--1031, 2019.

\bibitem[Vovk et~al.(2018)Vovk, Nouretdinov, Manokhin, and
  Gammerman]{vovk2018cross}
Vovk, V., Nouretdinov, I., Manokhin, V., and Gammerman, A.
\newblock Cross-conformal predictive distributions.
\newblock In \emph{The Journal of Machine Learning Research (JMLR)}, pp.\
  37--51, 2018.

\bibitem[Wager \& Athey(2018)Wager and Athey]{wager2018estimation}
Wager, S. and Athey, S.
\newblock Estimation and inference of heterogeneous treatment effects using
  random forests.
\newblock \emph{Journal of the American Statistical Association}, 113\penalty0
  (523):\penalty0 1228--1242, 2018.

\bibitem[Wager et~al.(2014)Wager, Hastie, and Efron]{wager2014confidence}
Wager, S., Hastie, T., and Efron, B.
\newblock Confidence intervals for random forests: The jackknife and the
  infinitesimal jackknife.
\newblock \emph{The Journal of Machine Learning Research (JMLR)}, 15\penalty0
  (1):\penalty0 1625--1651, 2014.

\bibitem[Welling \& Teh(2011)Welling and Teh]{welling2011bayesian}
Welling, M. and Teh, Y.~W.
\newblock Bayesian learning via stochastic gradient langevin dynamics.
\newblock In \emph{Proceedings of the 28th international conference on machine
  learning (ICML-11)}, pp.\  681--688, 2011.

\bibitem[White \& White(2010)White and White]{white2010interval}
White, M. and White, A.
\newblock Interval estimation for reinforcement-learning algorithms in
  continuous-state domains.
\newblock In \emph{Advances in Neural Information Processing Systems}, pp.\
  2433--2441, 2010.

\end{thebibliography}
\bibliographystyle{icml2020}

\onecolumn

\appendix

\setcounter{figure}{0}
\setcounter{table}{0}
\setcounter{equation}{0}

\section*{Appendix}

\section{Influence Functions: Background \& Key Concepts}
\label{AppendixA}
\subsection{Formal Definition}
\label{AppendixA1}
Robust statistics is the branch of statistics concerned with the detection of outlying observations. An estimator is deemed {\it robust} if it produces similar results as the majority of observations indicates, regardless of how a minority of other observations is perturbed (\cite{huber1981robust}). The influence function measures these effects in statistical functionals by analyzing the behavior of a functional not only at the distribution of interest, but also in an entire neighborhood of distributions around it. Lack of model robustness is a clear indicator of model uncertainty, and hence influence functions arise naturally in our method as a (pointwise) surrogate measure of model uncertainty. In this section we formally define influence functions and discuss its properties. 

The pioneering works in (\cite{hampel2011robust}) and (\cite{huber1981robust}) coined the notion of influence functions to assess the robustness of statistical functionals to perturbations in the underlying distributions. Consider a statistical functional $T: \mathcal{P} \to \mathbb{R}$, defined on a probability space $\mathcal{P}$, and a probability distribution $\mathbb{P} \in \mathcal{P}$. Consider distributions of the form $\mathbb{P}_{\varepsilon,z} = (1 - \varepsilon)\mathbb{P} + \varepsilon \Delta z$ where $\Delta z$ denotes the Dirac distribution in the point $z = (\boldsymbol{x}, y)$, representing the contaminated part of the data. For the functional $T$ to be considered robust, $T(\mathbb{P}_{\varepsilon,z})$ should not be too far away from $T(\mathbb{P})$ for any possible $z$ and any small $\varepsilon$. The limiting case of $\varepsilon \to 0$ defines the influence function. That is, Then the influence
function of $T$ at $\mathbb{P}$ in the point $z$ is defined as
\begin{equation}
\mathcal{I}(z; \mathbb{P}) = \lim_{\varepsilon \to 0} \frac{T(\mathbb{P}_{\varepsilon, z}) - T(\mathbb{P})}{\varepsilon} \triangleq \left. \frac{\partial}{\partial \varepsilon}\, T(\mathbb{P}_{\varepsilon, z})\, \right|_{\varepsilon = 0},
\label{eqA1}
\end{equation}
The influence function measures the robustness of $T$ by quantifying the effect on the estimator $T$ when adding an infinitesimally small amount of contamination at the point $z$. If the supremum of $\mathcal{I}(.)$ over $z$ is bounded, then an infinitesimally small amount of perturbation cannot cause arbitrary large changes in the estimate. Then small amounts of perturbation cannot completely change the estimate which ensures the robustness of the estimator.

\subsection{The von Mises Expansion}
\label{AppendixA2}
The von Mises expansion is a distributional analog of the Taylor expansion applied for a functional instead of a function. For two distributions $\mathbb{P}$ and $\mathbb{Q}$, the Von Mises expansion is (\cite{fernholz2012mises}):
\begin{equation}
T(\mathbb{Q}) = T(\mathbb{P}) + \int \mathcal{I}^{(1)}(z; \mathbb{P}) \, d(\mathbb{Q} - \mathbb{P}) + \frac{1}{2} \int \mathcal{I}^{(2)}(z; \mathbb{P}) \, d(\mathbb{Q} - \mathbb{P}) + \, \ldots,   
\label{eqA2}
\end{equation}
where $\mathcal{I}^{(k)}(z; \mathbb{P})$ is the $k^{th}$ order influence function. By setting $\mathbb{Q}$ to be a perturbed version of $\mathbb{P}$, i.e., $\mathbb{Q} = \mathbb{P}_{\varepsilon}$, the von Mises expansion at point $z$ reduces to:
\begin{equation}
T(\mathbb{P}_{\varepsilon, z}) = T(\mathbb{P}) + \varepsilon\, \mathcal{I}^{(1)}(z; \mathbb{P})  + \frac{\varepsilon^2}{2}\, \mathcal{I}^{(2)}(z; \mathbb{P}) \, + \, \ldots,   
\label{eqA3}
\end{equation} 
and so the $k^{th}$ order influence function is operationalized through the derivative
\begin{equation}
\mathcal{I}^{(k)}(z; \mathbb{P}) \triangleq \left. \frac{\partial}{\partial^k \varepsilon}\, T(\mathbb{P}_{\varepsilon, z})\, \right|_{\varepsilon = 0}.   
\label{eqA4}
\end{equation}

\subsection{Influence Function of Model Loss}
\label{AppendixA3}
Now we apply the mathematical definitions in Sections \ref{AppendixA1} and \ref{AppendixA2} to our learning setup. In our setting, the functional $T(.)$ corresponds to the (trained) model parameters $\hat{\theta}$ and the distribution $\mathbb{P}$. In this case, influence functions of $\hat{\theta}$ computes how much the model parameters would change if the underlying data distribution was perturbed infinitesimally.   
\begin{equation}
\mathcal{I}^{(1)}_\theta(z) = \left. \frac{\partial\, \hat{\theta}_{\varepsilon, z}}{\partial\, \varepsilon}\, \, \right|_{\varepsilon = 0},\,\,\, \hat{\theta}_{\varepsilon, z} \triangleq \arg \min_{\theta \in \Theta} \frac{1}{n} \sum_{i=1}^{n} \ell(z_i; \theta) + \varepsilon\,\ell(z; \theta).  
\label{eqA5}
\end{equation}
Recall that in the definition of the influence function $\mathbb{P}_{\varepsilon,z} = (1 - \varepsilon)\mathbb{P} + \varepsilon \Delta z$ where $\Delta z$ denotes the Dirac distribution in the point $z = (\boldsymbol{x}, y)$. Thus, the (first-order) influence function in (\ref{eqA5}) corresponds to perturbing a training data point $z$ by an infinitesimally small change $\varepsilon$ and evaluating the corresponding change in the learned model parameters $\hat{\theta}$. More generally, the $k^{th}$ order influence function of $\hat{\theta}$ is defined as follows:  
\begin{equation}
\mathcal{I}^{(k)}_\theta(z) = \left. \frac{\partial^k\, \hat{\theta}_{\varepsilon, z}}{\partial\, \varepsilon^k}\, \, \right|_{\varepsilon = 0}.   
\label{eqA6}
\end{equation}
By applying the von Mises expansion, we can approximate the parameter of a model trained on the training dataset with perturbed data point $z$ as follows: 
\begin{equation}
\hat{\theta}_{\varepsilon, z} \approx \hat{\theta} + \varepsilon\, \mathcal{I}_\theta^{(1)}(z)  + \frac{\varepsilon^2}{2}\, \mathcal{I}_\theta^{(2)}(z) \, + \, \ldots\, + \frac{\varepsilon^m}{m!}\, \mathcal{I}_\theta^{(m)}(z),   
\label{eqA7}
\end{equation} 
where $m$ is the number of terms included in the truncated expansion. When $m = \infty$, the exact parameter $\hat{\theta}_{\varepsilon, z}$ without the need to re-train the model. 

\subsection{Connection to leave-one-out estimators}
\label{AppendixA4}
Our uncertainty estimator depends on perturbing the model parameters by removing a single training point at a time. Note that removing a point $z$ is the same as perturbing $z$ by $\varepsilon = \frac{-1}{n}$,~hence~we~obtain~an ($m^{th}$ order) approximation of the parameter change due to removing the point $z$ as follows:
\begin{equation}
\hat{\theta}_{-z} - \hat{\theta} \approx \frac{-1}{n}\, \mathcal{I}^{(1)}_\theta(z) + \frac{1}{2n^2}\, \mathcal{I}^{(2)}_\theta(z) + \, \ldots\, + \frac{(-1)^m}{n^m \cdot m!}\, \mathcal{I}^{(m)}_\theta(z),
\label{eqA8}
\end{equation}
where $\hat{\theta}_{-z}$ is the model parameter learned by removing the data point $z$ from the training data.

\section{Derivation of Influence Functions}
Recall that the LOO parameter $\hat{\theta}_{i,\epsilon}$ is obtained by solving the optimization problem:
\begin{equation}
\hat{\theta}_{i,\epsilon} = \arg \min_{\theta \in \Theta} L(\mathcal{D}, \theta) + \epsilon \cdot \ell(y_i, f(\boldsymbol{x}_i; \theta)).
\label{eqB1}
\end{equation}
Let us first derive the first order influence function $\mathcal{I}^{(1)}(\boldsymbol{x}_i, y_i)$. Let us first define $\Delta_{i,\epsilon} \triangleq \hat{\theta}_{i,\epsilon} - \hat{\theta}$. The first order influence function is given by:
\begin{equation}
\mathcal{I}^{(1)}(\boldsymbol{x}_i, y_i) = \frac{\partial \hat{\theta}_{i,\epsilon}}{\partial \epsilon} = \frac{\partial \Delta_{i,\epsilon}}{\partial \epsilon}.
\label{eqB2}
\end{equation}
Note that, since $\hat{\theta}_{i,\epsilon}$ is the minimizer of (\ref{eqB1}), then the perturbed loss has to satisfy the following (first order) optimality condition: 
\begin{equation}
\nabla_\theta \left\{L(\mathcal{D}, \theta) + \epsilon \cdot \ell(y_i, f(\boldsymbol{x}_i; \theta))\right\}\big|_{\theta=\hat{\theta}_{i,\epsilon}} = 0.
\label{eqB3}
\end{equation}
Since $\lim_{\epsilon \to 0} \hat{\theta}_{i,\epsilon} = \hat{\theta}$, then we can write the following Taylor expansion:
\begin{equation}
\nabla_\theta \sum^{\infty}_{k=0} \frac{\Delta^k_{i,\epsilon}}{k!}\cdot \nabla^k_\theta  \left\{L(\mathcal{D}, \hat{\theta}) + \epsilon \cdot \ell(y_i, f(\boldsymbol{x}_i; \hat{\theta}))\right\}= 0.
\label{eqB4}
\end{equation}
Now by dropping the $o(\|\Delta_{i,\epsilon}\|)$ terms, we have:
\begin{equation}
\nabla_\theta \left( \left\{L(\mathcal{D}, \hat{\theta}) + \epsilon \cdot \ell(y_i, f(\boldsymbol{x}_i; \hat{\theta}))\right\} + \Delta_{i,\epsilon}\cdot \nabla_\theta \left\{L(\mathcal{D}, \hat{\theta}) + \epsilon \cdot \ell(y_i, f(\boldsymbol{x}_i; \hat{\theta}))\right\}\right)= 0.
\label{eqB5}
\end{equation}
Since $\hat{\theta}$ is a indeed a minimizer of the loss function $\ell(.)$, then we have $\nabla_\theta \ell(.) = 0$. Thus, (\ref{eqB5}) reduces to the following condition: 
\begin{equation}
\left\{\epsilon \cdot \nabla_\theta\,\,\ell(y_i, f(\boldsymbol{x}_i; \hat{\theta})) \right\} + \Delta_{i,\epsilon}\cdot \left\{ \nabla_\theta^2 \,\, L(\mathcal{D}, \hat{\theta}) + \epsilon \cdot \nabla^2_\theta \,\, \ell(y_i, f(\boldsymbol{x}_i; \hat{\theta}))\right\} = 0.
\label{eqB6}
\end{equation}
By solving for $\nabla_\theta$, we have
\begin{equation}
\Delta_{i,\epsilon} = -\left\{ \nabla_\theta^2 \,\, L(\mathcal{D}, \hat{\theta}) + \epsilon \cdot \nabla^2_\theta \,\, \ell(y_i, f(\boldsymbol{x}_i; \hat{\theta}))\right\}^{-1} \cdot \left\{\epsilon \cdot \nabla_\theta\,\, \ell(y_i, f(\boldsymbol{x}_i; \hat{\theta})) \right\},
\label{eqB7}
\end{equation}
which can be approximated by keeping only the $O(\epsilon)$ terms as follows:
\begin{equation}
\Delta_{i,\epsilon} = -\left\{ \nabla_\theta^2 \,\, L(\mathcal{D}, \hat{\theta}) \right\}^{-1} \cdot \left\{\epsilon \cdot \nabla_\theta\,\, \ell(y_i, f(\boldsymbol{x}_i; \hat{\theta})) \right\}.
\label{eqB8}
\end{equation}
Noting that $\nabla_\theta^2 \,\, L(\mathcal{D}, \hat{\theta})$ is the Hessian matrix $H_{\hat{\theta}}$, we have:
\begin{equation}
\Delta_{i,\epsilon} = -H_{\hat{\theta}}^{-1} \cdot \left\{\epsilon \cdot \nabla_\theta\,\, \ell(y_i, f(\boldsymbol{x}_i; \hat{\theta})) \right\}.
\label{eqB9}
\end{equation}
By taking the derivative with respect to $\epsilon$, we arrive at the expression for first order influence functions:
\begin{equation}
\mathcal{I}^{(1)}(\boldsymbol{x}_i, y_i) = \frac{\Delta_{i,\epsilon}}{\epsilon}\big|_{\epsilon = 0} = -H_{\hat{\theta}}^{-1} \cdot \nabla_\theta\,\, \ell(y_i, f(\boldsymbol{x}_i; \hat{\theta})).
\label{eqB10}
\end{equation}
Now let us examine the second order influence functions. In order to obtain $\mathcal{I}^{(2)}(\boldsymbol{x}_i, y_i)$, we need to differentiate (\ref{eqB6}) after omitting the $O(\epsilon)$ once again as follows:
\begin{equation}
\left\{2\Delta_{i,\epsilon}\cdot \epsilon \cdot \nabla^2_\theta\,\,\ell(y_i, f(\boldsymbol{x}_i; \hat{\theta})) \right\} + \left\{\Delta^2_{i,\epsilon}\cdot \nabla_\theta^2 \,\, L(\mathcal{D}, \hat{\theta}) + \Delta_{i,\epsilon}\cdot \Delta_{i,\epsilon}\cdot \nabla_\theta^3 \,\, L(\mathcal{D}, \hat{\theta})\right\} = 0.
\label{eqB11}
\end{equation}
Where we have applied the chain rule to obtain the above. By substituting $\nabla_\theta^2 \,\, L(\mathcal{D}, \hat{\theta}) = H_{\theta}$ and dividing both sides of (\ref{eqB11}) by $\epsilon^2$, we have
\begin{equation}
\left\{2\,\frac{\Delta_{i,\epsilon}}{\epsilon} \cdot \nabla^2_\theta\,\,\ell(y_i, f(\boldsymbol{x}_i; \hat{\theta})) \right\} + \left\{\frac{\Delta^2_{i,\epsilon}}{\epsilon^2}\cdot H_\theta + \left(\frac{\Delta_{i,\epsilon}}{\epsilon}\right)^2\cdot \nabla_\theta^3 \,\, L(\mathcal{D}, \hat{\theta})\right\} = 0.
\label{eqB12}
\end{equation}
Thus, by re-arranging (\ref{eqB11}), we can obtain $\mathcal{I}^{(2)}(\boldsymbol{x}_i, y_i)$ in terms of $\mathcal{I}^{(1)}(\boldsymbol{x}_i, y_i)$ as follows:
\begin{equation}
\mathcal{I}^{(2)}(\boldsymbol{x}_i, y_i) = - H^{-1}_\theta\,\left(\left(\mathcal{I}^{(1)}(\boldsymbol{x}_i, y_i)\right)^2 \cdot \nabla_\theta^3 \,\, L(\mathcal{D}, \hat{\theta}) + 2\, \mathcal{I}^{(1)}(\boldsymbol{x}_i, y_i)\cdot \nabla^2_\theta\,\,\ell(y_i, f(\boldsymbol{x}_i; \hat{\theta}))\right). \nonumber
\end{equation}

Similarly, we can obtain the $k^{th}$ order influence function, for any $k>1$, by repeatedly differentiating equation (\ref{eqB6}) $k$ times, i.e.,
\begin{equation}
\frac{\partial}{\partial \epsilon^k}\left\{\epsilon \cdot \nabla_\theta\,\,\ell(y_i, f(\boldsymbol{x}_i; \hat{\theta})) + \Delta_{i,\epsilon}\cdot \nabla_\theta^2 \,\, L(\mathcal{D}, \hat{\theta}) \right\} = 0.
\label{eqB13}
\end{equation}
and solving for $\partial \Delta^k_{i,\epsilon}/ \partial \epsilon^k$. By applying the higher-order chain rule to (\ref{eqB13}) (or equivalently, take the derivative of $\mathcal{I}^{(2)}(\boldsymbol{x}_i, y_i)$ for $k-2$ times), we recover the expressions in Definition 2 and Lemma 3 in \cite{giordano2019higher}. 
\newpage
\section{Theorem 1}
Theorem 1 follows from Theorem 1 in \cite{barber2019predictive} for $m \to \infty$ when all HOIFs exist.

Recall that the exact DJ interval width is bounded above by:
\begin{align}
W(\widehat{\mathcal{C}}^{(\infty)}_{\alpha, n}(\boldsymbol{x};\hat{\theta})) \leq 2\, \widehat{Q}_{\alpha, n}(\mathcal{R}_n) + 2\, \widehat{Q}_{\alpha, n}(\mathcal{V}_{n}(\boldsymbol{x})).
\label{eqC4}
\end{align} 
Since the term $\widehat{Q}_{\alpha, n}(\mathcal{R}_n)$ is constant for any $\boldsymbol{x}$, discrimination boils down to the following condition:
\begin{align}
\mathbb{E}[\,\widehat{Q}_{\alpha, n}(\mathcal{V}_{n}(\boldsymbol{x}))\,] \, \geq \, \mathbb{E}[\,\widehat{Q}_{\alpha, n}(\mathcal{V}_{n}(\boldsymbol{x}^{\prime}))\,] \Leftrightarrow \mathbb{E}[\,\ell(y, f(\boldsymbol{x};\hat{\theta}))\,] \, \geq \, \mathbb{E}[\,\ell(y^{\prime}, f(\boldsymbol{x}^{\prime};\hat{\theta}))\,].
\label{eqC4}
\end{align}
Note that to prove the above, it suffices to prove the following:
\begin{align}
\mathbb{E}[\,v_i(\boldsymbol{x})\,] \, \geq \, \mathbb{E}[\,v_i(\boldsymbol{x}^\prime)\,] \Leftrightarrow \mathbb{E}[\,\ell(y, f(\boldsymbol{x};\hat{\theta}))\,] \, \geq \, \mathbb{E}[\,\ell(y^{\prime}, f(\boldsymbol{x}^{\prime};\hat{\theta}))\,].
\label{eqC5}
\end{align}

If the model is stable (based on the definition in \cite{bousquet2002stability}), then a classical result by \cite{devroye1979distribution} states that:
\begin{align}
\mathbb{E}[\,|\ell(y, f(\boldsymbol{x};\hat{\theta})) - \ell_n(y,f(\boldsymbol{x};\hat{\theta}))|^2\,]  \approx \, \mathbb{E}[\,|\ell(y, f(\boldsymbol{x};\hat{\theta})) - \ell(y,f(\boldsymbol{x};\hat{\theta}_{-i}))|^2\,] + Const., 
\label{eqC6}
\end{align}
as $n \to \infty$, where $\ell_n(.)$ is the empirical risk on the training sample, and the expectation above is taken over $y\,|\,\boldsymbol{x}$. From (\ref{eqC6}), we can see that an increase in the LOO risk $\ell(y,f(\boldsymbol{x};\hat{\theta}_{-i}))$ implies an increase in the empirical risk $\ell_n(y,f(\boldsymbol{x};\hat{\theta}))$, and vice versa. Thus, for any two feature points $\boldsymbol{x}$ and $\boldsymbol{x}^\prime$, if $v(\boldsymbol{x})$ is greater than $v(\boldsymbol{x}^\prime)$, then on average, the empirical risk at $\boldsymbol{x}$ is greater than that at $\boldsymbol{x}^\prime$.  

\section{Experimental Details}

\subsection{Implementation of Baselines}
In what follows, we provide details for the implementation and hyper-parameter settings for all baseline methods involved in Section \ref{Sec5}. 

{\bf Probabilistic backpropagation (PBP).} We implemented the PBP method proposed in (\cite{hernandez2015probabilistic}) with inference via expectation propagation using the \texttt{theano} code provided by the authors in (\texttt{github.com/HIPS/Probabilistic-Backpropagation}).~Training~was conducted via 1000 epochs.  

{\bf Monte Carlo Dropout (MCDP).} We implemented a \texttt{Pytorch} version of the MCDP method proposed in (\cite{gal2016dropout}). In all experiments, we tuned the dropout probability using Bayesian optimization to optimize the AUC-ROC performance on the training sample. We used 1000 samples at inference time to compute the mean and variance of the predictions. The credible intervals were constructed as the $(1-\alpha)$ quantile function of a posterior Gaussian distribution defined by the predicted mean and variance estimated through the Monte Carlo outputs. Similar to the other baselines, we conducted training via 1000 epochs for the SGD algorithm.

{\bf Bayesian neural networks (BNN).} We implemented BNNs with inference via stochastic gradient Langevin dynamics (SGLD) (\cite{welling2011bayesian}). We initialized the prior weights and biases through a uniform distribution over $[-0.01, 0.01]$. We run 1000 epochs of the SGLD inference procedure and collect the posterior distributions to construct the credible intervals.

{\bf Deep ensembles (DE).} We implemented a \texttt{Pytorch} version of the DE metho (without adversarial training)d proposed in (\cite{lakshminarayanan2017simple}). We used the number of ensemble members $M=5$ as recommended in the recent study in (\cite{ovadia2019can}). Predictions of the different ensembles were averaged and the confidence interval was estimated as 1.645 multiplied by the empirical variance for a target coverage of 90$\%$. We trained the model through 1000 epochs.

\end{document}